\RequirePackage{tikz}
\documentclass[sn-apacite]{sn-jnl}

\usetikzlibrary{positioning,arrows,calc,math,angles,quotes,patterns,shapes.geometric,matrix}
\usetikzlibrary{arrows.meta,babel}
\usepackage{caption}
\usepackage{hyperref}
\usepackage[utf8]{inputenc}
\usepackage[T1]{fontenc}
\usepackage{physics}
\usepackage{lineno}
\usepackage{natbib}
\usepackage{diagbox}
\setcitestyle{authoryear,open={(},close={)}}
\jyear{2023}%

\theoremstyle{thmstyleone}%
\theoremstyle{thmstyletwo}%

\theoremstyle{thmstylethree}%

\begin{document}

\title[Comparison of Metaheuristics for the Firebreak Placement Problem]{Comparison of metaheuristics for the firebreak placement problem: a simulation-based optimization approach}


\author[1,4]{\fnm{David} \sur{Palacios-Meneses}}

\author*[2,4]{\fnm{Jaime} \sur{Carrasco}} \email{jcarrascob@utem.cl}

\author[3]{\fnm{Sebastián} \sur{Dávila-Gálvez}}

\author[1]{\fnm{Maximiliano} \sur{Martínez}}

\author[1,4]{\fnm{Rodrigo} \sur{Mahaluf}}
\author[1,4]{\fnm{Andrés} \sur{Weintraub}}

\affil[1]{\orgdiv{Industrial Engineering Department}, \orgname{University of Chile}, \orgaddress{\city{Santiago}, \country{Chile}}}

\affil[2]{\orgdiv{Departamento de Industria, Facultad de Ingenier\'ia}, \orgname{Universidad Tecnol\'ogica Metropolitana},  \orgaddress{\city{Santiago}, \country{Chile}}}

\affil[3]{\orgdiv{Industrial Engineering Department, Laboratory for the Development of Sustainable Production Systems (LDSPS)}, \orgname{Universidad de Santiago de Chile},  \orgaddress{\city{Santiago}, \country{Chile}}}

\affil[4]{\orgname{Complex Engineering System Institute - ISCI}, \orgaddress{\city{Santiago}, \country{Chile}}}

\abstract{The problem of firebreak placement  is crucial for fire prevention, and its effectiveness at landscape scale will depend on their ability to impede the progress of future wildfires. To provide an adequate response, it is therefore necessary to consider the stochastic nature of fires, which are highly unpredictable from ignition to extinction. Thus, the placement of firebreaks can be considered a stochastic optimization problem where: (1) the objective function is to minimize the expected cells burnt of the landscape; (2) the decision variables being the location of firebreaks; and (3) the random variable being the spatial propagation/behavior of fires. In this paper, we propose a solution approach for the problem from the perspective of simulation-based optimization (SbO), where the objective function is not available (a black-box function), but can be computed (and/or approximated) by wildfire simulations. For this purpose, Genetic Algorithm and GRASP are implemented. The final implementation yielded favorable results for the Genetic Algorithm, demonstrating strong performance in scenarios with medium to high operational capacity, as well as medium levels of stochasticity.}




\maketitle

\section{Introduction}\label{sec1}

Wildfires have intensified and become a significant cause for concern at both the national and global levels, owing to the serious harm inflicted upon our ecosystems and human communities \citep{kelly2020fire,pausas2021wildfires}. How can we effectively address this critical problem, which recurs every summer season? There are two established approaches for managing wildfires, primarily differing in their application timing: preventive or reactive. The former, involves actions taken on the landscape before fires occur to reduce the fire's rate of spread, intensity, and flame length, as well as to prevent crowning and spot fire development \citep{amiro2001fire,finney2001design}. These activities, known as \textit{forest fuel management} (FFM), typically involve vegetation reduction and manipulation, fuel conversion, or isolation, such as firebreaks \citep{agee2005basic}. The latter, known as firefighting or fire suppression, includes all activities aimed at controlling and ultimately extinguishing a wildfire after ignition. In general, firefighting is the most popular approach because it is carried out in real-time, requires a significant allocation of resources, and is highly visible in the media during the summer season, especially when fires escape suppression efforts or occur near the wildland-urban interface \citep{martell2007forest}. However, in light of recent global events, such as the 2004 Alaska fire season, the 2019/2020 Australian bushfires, the 2021 Dixie Fire in California, and Canada's ongoing 2023 fire season, which is on track to be one of the most devastating in the country's history, it appears that reactive measures alone are not enough.

The issue of FFM has been tackled through Operations Research, by introducing the risk of fire and uncertainty primarily as parameters of mixed-integer programming models (MIPs). However, the application of these models to the field of wildfire management is relatively recent \citep{martell2007forest}. Several studies have integrated elementary fire ignition and propagation process models into spatially explicit integer and dynamic programming models (see, e.g., \cite{bettinger2009,carrasco2023firebreak,kim2009spatial,konoshima2008spatial,acuna2010integrated,gonzalez2011integrating}). Other more sophisticated approaches have incorporated fire uncertainty into the FFM through the use of Stochastic Linear Programming (SLP). In fact, a variety of parameters can be randomized or incorporated as random variables to account for the inherent uncertainty in fire behavior and weather conditions. For example, fuel moisture content, treatment costs, weather variables such as temperature, humidity, wind speed and direction, precipitation, and atmospheric stability could be considered stochastic parameters. \cite{boychuk1996multistage} developed a multistage stochastic programming model for sustainable timber supply at the forest level incorporating fire risk. The model involves using probability distributions to model the likelihood and potential impact of fire events, as well as the potential effectiveness of different fire management strategies. However, the paper focuses on a hypothetical forest and simplified assumptions of the parameters that account for fire risk. \cite{kabli2015stochastic} proposes a two-stage stochastic programming model to optimize fuel treatment decisions in forest management. Specifically, the authors use probabilistic distributions to represent the uncertainty in fire ignition and spread. These distributions are incorporated into the model as stochastic parameters, which are then used to generate a range of possible scenarios for fire risk and treatment effectiveness.

A widely used parameter is known as burn probability (BP), which is usually obtained using a procedure that involves an iterative and concatenated process of simulating spatially explicit fire ignition and fire growth scenarios across the landscape \citep{parisien2005mapping,finney2005challenge}.  Specifically, BP is calculated as the percentage of simulations in which the fire reaches a particular point or area \citep{parisien2005mapping}. Wildfire simulation models such as FARSITE \citep{finney1998farsite}, Burn-P3 \citep{parisien2005mapping} and Cell2Fire \citep{pais2021cell2fire} have the ability to generate fire spread simulations and calculate burn probability. BP models have been used to support fuel reduction strategies, such as prescribed burns and firebreaks, by identifying areas where fuel reduction efforts will have the greatest impact in reducing burn probability \citep{carrasco2023firebreak,ager2010comparison,oliveira2016assessing}.

The design of effective firebreak placement at landscape scale presents an interesting challenge for decision support modeling and is considered an open problem within the Natural Resources and Operations Research field in Forestry \citep{martell2007forest,ronnqvist2015operations}. Specifically, the present problem has not been sufficiently addressed, mainly due to the following three reasons: (i) unrealistic (fictitious and small-scale) forest landscapes have been used; (ii) models have been developed with simplified fire spread simulators, which allow numerous simulations over large areas, but do not consider systems of fire behavior (see, e.g, \cite{bettinger2009, kim2009spatial, konoshima2008spatial, gonzalez2011integrating}); iii) a measure of the effectiveness of the spatial interaction between fuel management and wildfire has not been considered.   

Simulation-based optimization (SbO) is an optimization approach that uses computer simulation models to evaluate the performance of various decision variables with the goal of finding the optimal solution that maximizes or minimizes a function that is not directly available but can be computed by simulation \citep{gosavi2015simulation}. Compared to traditional stochastic optimization techniques, one of the main advantages of SbO is its ability to handle complex, nonlinear, and stochastic systems in which the relationships between decision variables and the objective function are difficult to define in closed-form mathematical terms due to inherent system uncertainty or lack of knowledge. Thus, this is particularly useful in fields such as engineering and/or natural systems, where real-world problems are often too complex to be modeled analytically, especially when the problem is highly nonlinear or involves complex interactions (see, e.g., \cite{gaur2011analytic,khalili2022stochastic,rani2010simulation,RytwinCrowe}).

The effectiveness of the firebreak placement at the landscape scale depends on their ability to impede the progress of future wildfires. Consequently, a realistic solution should include the interaction of firebreaks with potential fires that may spread across the landscape, as well as a measure of the effectiveness of the applied management to provide more robust and reliable solutions. In the present study, we hypothesize that an effective firebreak placement can be constructed under a simulation-based optimization approach in a reasonable performance time, where the objective function can be formulated as a black-box function that can be computed through wildfire simulations.

\section{Theoretical background}\label{sec2}
In this section, it is formally given some basic concepts and notions from Simulation-based Optimization (SbO) and the Firebreak Placement Problem (FPP) in order to provide the readers with all the necessary background to follow this paper.

In this formulation, we assume that the landscape is divided into a set of cells denoted by $\mathcal{N}$, and we denote the binary decision variables by a vector $\vb{x} \in \{0,1\}^{\lvert\mathcal{N}\vert} $, where $x_{j}$ equals 1 if the cell $j$ is selected to place a firebreak, and 0 otherwise. The application of a firebreak to a landscape cell implies the total removal of the vegetation fuel at that location, and therefore the firebreaks are non-flammable. Due to the high costs involved in constructing firebreaks and, broadly, fuel treatments, in practice, only a small percentage of the forest is managed \citep{oliveira2016assessing, jingan2005land}. In our study, we consider a range of percentage that will not exceed 15\% of the total flammable cells in the landscape. For this reason, we introduce a parameter $\alpha \in [0, 1]$, which restricts the feasible space. Therefore, we formalize the feasible region (constraints set) of the optimization problem as follows:
\begin{equation} 
    \Omega_\alpha := \{ \vb{x} \in \{0,1\}^{\lvert \mathcal{N}\vert}: \sum_{j\in\mathcal{N}}x_{j}\leq\alpha\lvert \mathcal{N}\vert  \}
    \label{Eq:Constraint}
\end{equation}
Our experiments set the $\alpha$ values 0.01, 0.03, 0.05, 0.075, 0.1, and 0.15 (or 1\%, 3\%, 5\%, 7.5\%, 10\%, and 15\%, respectively). We denote by $I_\alpha(x) \subseteq \mathcal{N}$ the set of indices associated with the feasible point $x \in \Omega_\alpha$. 

\begin{figure}[ht]
    \centering
\includegraphics[width=1.0\textwidth]{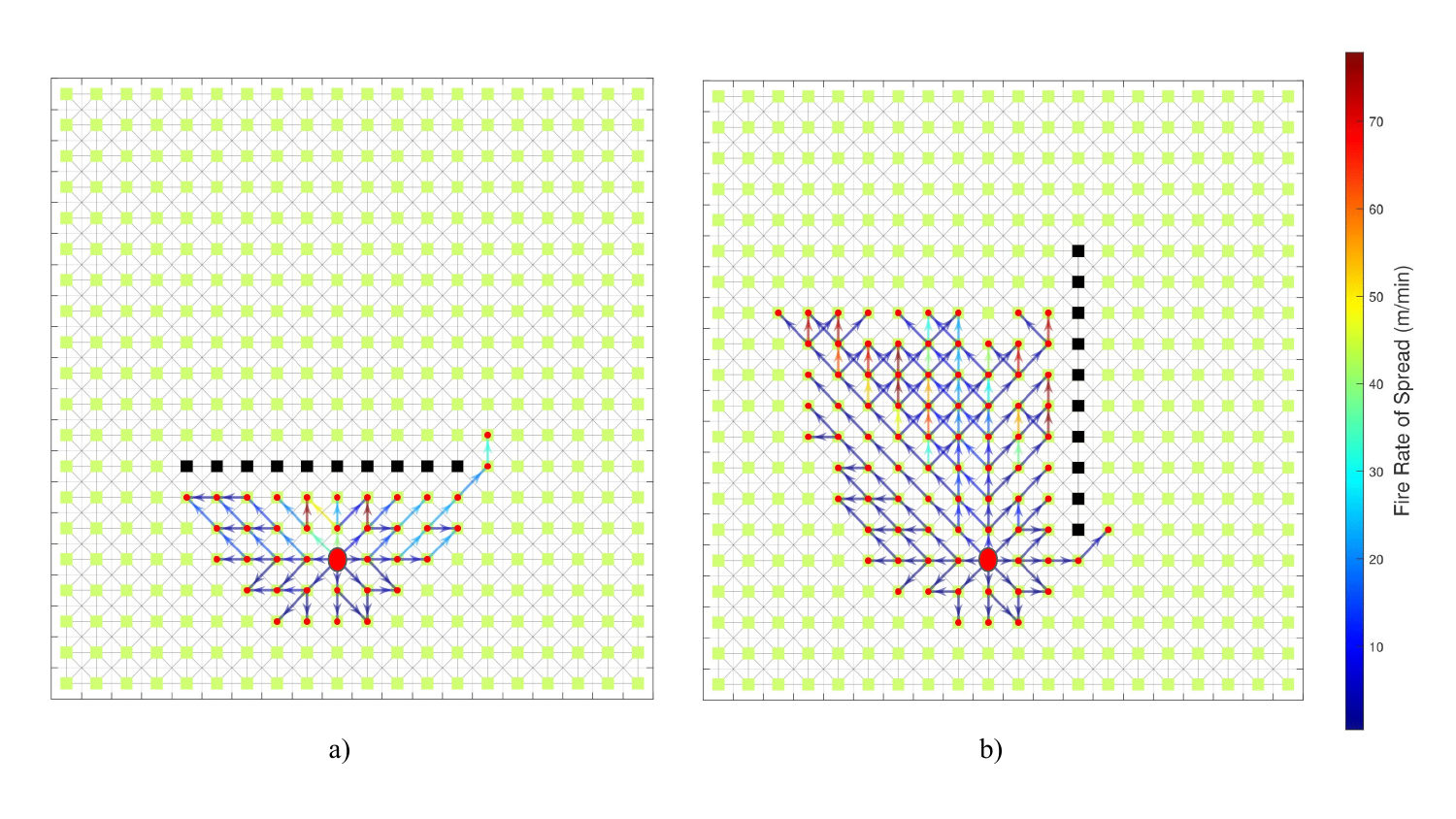}
\caption{Firebreak efficiency. a) and b) represent an efficient and an inefficient firebreak configuration, respectively. The red ellipsis indicates the ignition point.}
\label{fig:firebreak_efficiency}
\end{figure}

Here, we want to solve the following optimization problem:
\begin{equation} \label{Eq:FMP}
    \min_{\vb{x} \in \Omega_\alpha} \varphi(\vb{x}):=\mathbb{E}_\mathcal{\xi} \left[ L \left(\vb{x},\xi \right) \right]
\end{equation}
where $ L\left(\vb{x},\xi \right) $ is the random variable that represents the loss due to a random fire $\xi$ on the landscape $\mathcal{F}$, given the firebreak allocation decisions variables $\vb{x}$. The decision $\vb{x}$ will be as efficient as their ability to intersect the fire ($\xi$), and therefore hinder its spread (see Fig.\:\ref{fig:firebreak_efficiency}). The term ``simulation-based optimization (SbO)'' or ``simulation-optimization''  generally refers to optimization in the setting where the objective function  $\varphi(\vb{x})$ cannot be computed exactly, and its estimation depends on the realization of the random variables of the system. Thus, $\varphi(\vb{x})$ is not directly available; rather, realizations of $\xi$ come from the observed simulation replication outputs.

Therefore, the function $\varphi(\vb{x})$ is the expected value of a random variable $ L\left(\vb{x},\xi \right) $ where $\xi$ represents the randomness in our study: the fire ignition and fire spread in the landscape. The distribution of $ L\left(\vb{x},\xi \right) $ is an unknown function of the vector of decision variables $\vb{x}$, but realizations of $ L\left(\vb{x},\xi \right) $ can be observed through simulation experiments. In this study, we use a simulator known as Cell2Fire \citep{pais2021cell2fire}, which models the fire spreading as a cellular automata modelling the landscape as a regular grid, composed by cells characterized by a set of environmental variables including fuel type and topographic features. Cell2Fire effectively integrates decision making with spatial simulation - an advantage that sets it apart from simulators such as Prometheus \citep{tymstra2010development} or Burn-P3 \citep{parisien2005mapping}. With this software, we can simulate a set $R$ of independent and identically distributed (i.i.d.) replications, $L(\vb{x},\xi_1),L(\vb{x},\xi_2),...,L(\vb{x},\xi_R)$ at any $\vb{x}$. Thus, we can approximate the function $\varphi(\vb{x})$ by:
\begin{equation} \label{Eq:EVAL}
    \Bar{\varphi}(\vb{x}) = \dfrac{1}{R} \sum_{r=1}^{R}L(\vb{x},\xi_r)
\end{equation}
where the sampling errors can be controlled by increasing the number of i.i.d. replications $R$.

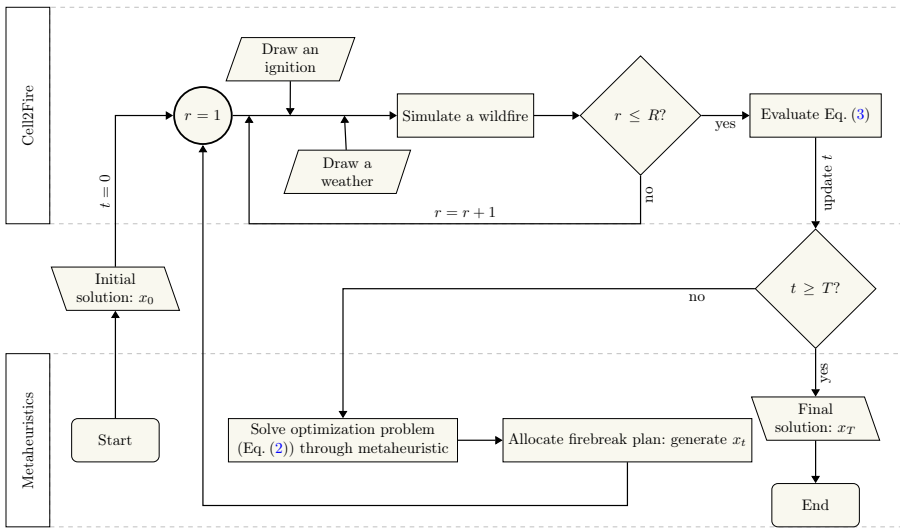
\begin{figure} [H]
    \centering
\usetikzlibrary{patterns}
\resizebox{\textwidth}{!}{
\begin{tikzpicture}
    \tikzstyle{box}=[rectangle,draw,node distance=1cm,text width=1cm,text centered,rounded corners,minimum height=2em,thick],
\tikzstyle{decision} = [diamond, minimum width=2cm, minimum height=1cm, text centered, text width=2cm, draw=black, fill=olive!5]
\tikzstyle{io} = [trapezium, trapezium left angle=70, trapezium right angle=110, minimum
width=1cm,   
text width=2cm,
trapezium stretches=true,
 minimum height=1cm, text centered, draw=black, fill=olive!5]

\tikzstyle{parameter}=[circle, draw=black, fill=olive!5, very thick, minimum size=1cm, text width=1cm, text centered]
\tikzstyle{startstop} = [rectangle, rounded corners, minimum width=2cm, minimum height=1cm,text centered, draw=black, fill=olive!5]

\tikzstyle{process} = [rectangle, minimum width=3cm, minimum height=1cm, text centered, draw=black, fill=olive!5]
\tikzstyle{operation} = [rectangle, minimum width=3cm, minimum height=1cm, text centered, draw=black, fill=olive!5]

\tikzstyle{output} = [trapezium, trapezium left angle=110, trapezium right angle=70, minimum
width=1cm,   
text width=2cm,
trapezium stretches=true,
 minimum height=1cm, text centered, draw=black, fill=olive!5]

 \tikzstyle{opt_problem_box} = [trapezium, trapezium left angle=90, trapezium right angle=90, minimum
width=1cm,   
text width=5cm,
trapezium stretches=true,
 minimum height=1cm, text centered, draw=black, fill=olive!5]

\tikzstyle{arrow} = [thick,->,>=Triangle]

\pgfdeclarelayer{back}    
\pgfsetlayers{back,main} 


\node (c2f)[rectangle,draw = black,rotate=90,    minimum width = 5cm, 
    minimum height = 1cm] {Cell2Fire};

\node (meta) [rectangle,draw = black,below of=c2f,yshift=-6.5cm,rotate=90,minimum width = 4cm, 
    minimum height = 1cm] {Metaheuristics};

\node (start) [startstop,right of= meta, xshift=1cm] {Start};

\node (initsol) [output,above of=start, yshift=2.5cm] {Initial solution: $x_0$};

\node (initsim) [parameter,right of=c2f , xshift=3cm] {$r=1$};

\node (wildfire) [process,right of=initsim , xshift=5cm] {Simulate a wildfire};

\node (weather) [io,above of=initsim , xshift=2cm,yshift=0.3cm] {Draw an ignition};

\node (ignition) [io,below of=wildfire , xshift=-2.7cm,yshift=-0.3cm] {Draw a weather};

\node (sr) [decision,right of=wildfire , xshift=3cm] {$r\leq R?$}; 

\node (mc) [operation,right of=sr , xshift=3cm] {Evaluate Eq.\:(\ref{Eq:EVAL})};

\node (tt) [decision,below of=mc, yshift=-3cm] {$t\geq T?$}; 

\node (solvemin) [opt_problem_box,right of=meta , xshift=6.2cm] {Solve optimization problem (Eq.\:(\ref{Eq:FMP})) through metaheuristic};

\node (allocate) [operation,right of=solvemin , xshift=5.5cm] {Allocate firebreak plan: generate $x_t$};

\node (optsol) [output,below of=tt, yshift=-2cm] {Final solution: $x_T$};

\node (end) [startstop,below of=optsol, yshift=-1cm] {End}; 

\draw [arrow] (start) -- (initsol);

\draw [arrow] (initsol) -- node[pos=0.5,anchor=south,rotate=90] {$t=0$} ++(0,4)  |-  (initsim.west) ;

\draw [arrow] (initsim) -- 
coordinate[pos=0.1](repsim)
coordinate[pos=0.35](midweather) 
coordinate[pos=0.68](midignition) (wildfire);
\draw [arrow] (weather) -- (midweather);
\draw [arrow] (ignition) -- (midignition);
\draw [arrow] (wildfire) -- (sr);
\draw [arrow] (sr) |- node[pos=0.2,anchor=north,rotate=90] {no} ++(0,-2.5) -- ++(-8,0) node[pos=0.5,anchor=south] {$r=r+1$} -|(repsim);
\draw [arrow] (sr) -- node[anchor=north] {yes} (mc);
\draw [arrow] (mc) -- node[anchor=north,rotate=90] {update $t$} (tt);
\draw [arrow] (tt) -- node[pos=0.2,anchor=north] {no}  ++(-8,0)
-| (solvemin);
\draw [arrow] (solvemin) -- (allocate);

\draw [arrow] (allocate.south) -- ++(0,-1) -| (initsim);
\draw [arrow] (tt) -- node[anchor=north,rotate=90] {yes} (optsol);
\draw [arrow] (optsol) -- (end);

    \begin{pgfonlayer}{back}   
\node[rectangle,
    draw = lightgray,
    dashed,
    minimum width = 19.5cm, 
    minimum height = 5cm]  (r1) at (10.25,0) {};
\node[rectangle,
    draw = lightgray,
    dashed,
    minimum width = 19.5cm, 
    minimum height = 4cm] (r2) at (10.25,-7.5)  {} ;
\end{pgfonlayer}

  \end{tikzpicture}}
   \caption[Solution Scheme]{Solution scheme.}
   \label{fig:workScheme}
\end{figure}

Figure\:\ref{fig:workScheme} illustrates the concept of the process simulation (SbO) that we will conduct using metaheuristics and fire spread simulations in order to solve the FPP (Eq.\:(\ref{Eq:FMP})). In this iterative process, the first step is to select an initial feasible solution configuration ($\vb{x}^0$). Each iteration chooses a random ignition point and weather scenario, and the simulator is used to model fire spread and evaluate the solution. Based on this evaluation, the metaheuristic generates a new solution, which involves making local changes, such as adding a new treatment (if possible) or moving an existing treatment to another position. The specific local move is determined by the metaheuristic being used. The process continues until a stopping criterion is met, such as reaching a maximum number of iterations.

\begin{figure}[ht]
    \centering
\includegraphics[width=1.1\textwidth]{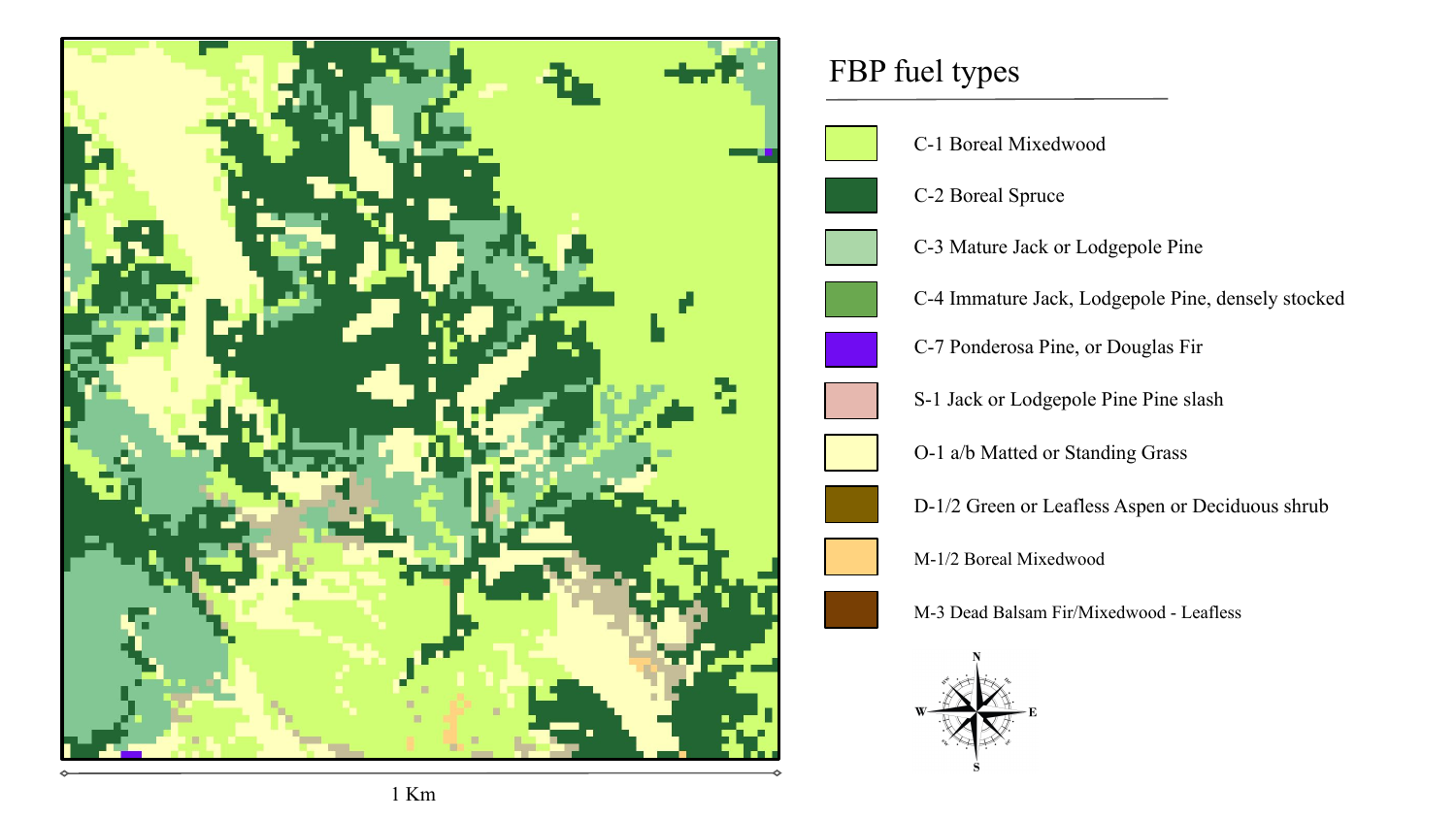}
    \caption{Study Area.}
    \label{Fig:StudyArea}
\end{figure}

To test our proposed solution approach, we conducted experiments using a real patch of a heterogeneous Canadian forest in the Alberta region. This landscape comprises a 10,000-hectare patch divided into $100\times 100$ square meter cells (100 m resolution).  Fig.\:\ref{Fig:StudyArea} depicts an overview of the study area.

The landscape data used in this study are consistent with the data used in the Prometheus software \citep{tymstra2010development} and/or the Cell2Fire system \citep{pais2021cell2fire}. These data include four raster layers representing landscape fuel models, elevation, slope, and aspect, as well as a lookup-table that associates each integer value in the fuel-type grid with a specific fuel type or non-fuel type.

\section{Metaheuristics application}
Metaheuristics are a high-level, problem-independent algorithmic framework that provides a set of guidelines or strategies for developing heuristic optimization algorithms \citep{sorensenmeta}. They have been shown to be effective for solving wildfire-related problems \citep{mendes2023iterated,jaafari2019genetic,mendez2016increase}, including the firebreak placement problem \citep{RytwinCrowe,Chung}.
In this study, we use two metaheuristics, Genetic Algorithm (GA) \citep{holland1992adaptation} and Greedy Randomized Adaptive Search Procedure (GRASP) \citep{feo1989probabilistic,feo1995greedy}, to solve the problem shown in Eq.\:(\ref{Eq:FMP}) using the structure presented in Figure Fig.\:\ref{fig:workScheme}. Some other metaheuristics were tested, such as Tabu Search, Artificial Bee Colony Algorithm and Ant Colony Optimization, but their results were not good enough to solve the problem. One central concept in metaheuristics is diversification \citep{GloverS19}, which is the process of exploring the solution space as much as possible. In order to ensure sufficient diversification, we test our solutions with a small value of $R$ (see Eq.\:(\ref{Eq:EVAL})). This value is a trade-off between quicker results with solutions that have high standard deviation (or noisy solutions) and slower results with solutions that have lower standard deviation, considering the Eq.\:(\ref{Eq:FMP}).

In the next section, we will illustrate some of the most important concepts of GA and GRASP.

\subsection{Algorithms}

\begin{itemize}
    
\item \textit{GRASP}: Greedy Randomized Adaptive Search Procedure (GRASP) \citep{feo1989probabilistic,feo1995greedy} is a greedy local search-based heuristic that constructs solutions in a two-stage procedure: a construction phase and a local search phase. In the first stage, an empty or partially constructed firebreak configuration is filled through a greedy methodology. Here, different candidates are inserted into the solution according to their improvement in the objective function of Eq.\:(\ref{Eq:FMP}). These candidates are selected according to user-defined greedy criteria and form a restricted candidate list (RCL). In our implementation, the greedy criteria ranks as better candidates the zones with a high frequency of burn. In the second phase of the algorithm, a local search procedure is executed around the solution obtained in the first stage, exploring different neighboring states with a similar greedy criteria. At the end of each iteration, the best firebreak configuration is stored, and the global solution is returned at the end of the run.
A illustrative example of a solution construction can be seen in Fig.\:\ref{fig:evolutiongrasp} and Fig.\:\ref{fig:graspall} depicts GRASP's working.
    
\item \textit{Genetic Algorithm}: Genetic Algorithms (henceforth GA) \citep{mirjalili2019genetic} are a family of metaheuristics that utilize the principles of The Theory of Evolution in their working. The main concepts that are needed to specify in a GA are four operations: Evaluation, Selection, Crossover, and Mutation. The following is a more detailed explanation of these, in addition to some previously required concepts and their application to our problem: 
    \begin{itemize}
    \item[-] \underline{Individual}: A firebreak configuration.
    
    \item[-] \underline{Population}: A set of individuals.
    
    \item[-] \underline{Generation}: The population of a specific iteration of the algorithm.
    
    \item[-] \underline{Offspring}: New individuals that will populate the next generation.

    \item[-] \underline{Selection}: Some population individuals are selected to generate offspring according to the value found in the evaluation phase. Generally, better individuals are selected, but there are other criteria for selection. In our implementation, we choose the half of the population with better performance for the next generation. This allows us to maintain the best configurations but keep the chance to include some bad solutions which may improve in the future.
    
    \item[-] \underline{Evaluation}: The solutions of the population are evaluated according to Eq.\:(\ref{Eq:FMP}). 

    \item[-] \underline{Crossover}: In this stage, some population solutions are combined to generate new individuals for the next generation. In our implementation, this stage generates the second half of the next generation (the first half is obtained in the selection). Here, we randomly pick two individuals selected in the Selection stage and combine them randomly in order to generate new offspring. This process is repeated until the number of offspring needed is fulfilled.
    
    \item[-] \underline{Mutation}: Some offspring generated are modified to diversify the solutions.
    
    \end{itemize}
    
\end{itemize}    

The cycle of a generation can be seen in Fig.\:\ref{fig:evolutionga}, while Fig.\:\ref{fig:allga} provides an overview of the main workings of the GA.

\subsection{Comparison Procedure}\label{seccomp}
In order to compare the metaheuristics outlined above, we will measure their performance in terms of the burn frequency of the cells (according to Eq.\:(\ref{Eq:FMP})). The comparison procedure considers five different random seeded tests for each metaheuristic. Each test is run for two hours, and their evaluation consists on a run of 1,000 simulations (i.e., $R=1,000$, see Eq.\:(\ref{Eq:EVAL})) with each optimal solution.

For results visualization, we will use boxplots and line dispersion plots. The boxplots will be built using the final solutions of the algorithms. Therefore, each box will represent the distribution of the results of the 5 different instances. Every instance will be compared against a random and a greedy solution with the same number of firebreak cells as in its respective case. The random solution consists on a random allocation of firebreaks, while the greedy solutions are constructed by simulate wildfires with no firebreaks, and then allocating firebreaks in the most burned cells.

The line dispersion plots will use the solutions obtained along the runtime of the metaheuristics, and they will show the evolution of the current best solution of the algorithm. It should be noted that this solution is obtained with a small value of $R$, so it is a noisy solution and may vary from the values shown in the boxplots.
Solutions are obtained using an Intel\textsuperscript{\textregistered} Core\textsuperscript{TM}  i7-9700K processor with 48GB of RAM. The metaheuristics are coded from scratch in Python 3.7.12, while the Cell2Fire simulator runs in C++ through Ubuntu S.O. version 18.04.

\subsection{Firebreak Pattern Assessment} \label{secpattern}
\begin{figure}[!h]
    \begin{minipage}[t]{0.5\textwidth}
        \centering
        \includegraphics[width=0.6\textwidth]{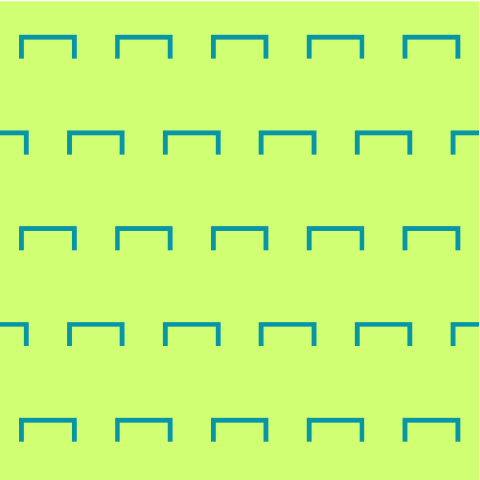}
        \caption{Firebreak Configuration.}
        \label{fig:fconfig}
    \end{minipage}
    \hfill
    \begin{minipage}[t]{0.5\textwidth}
        \centering
    \includegraphics[width=0.6\textwidth]{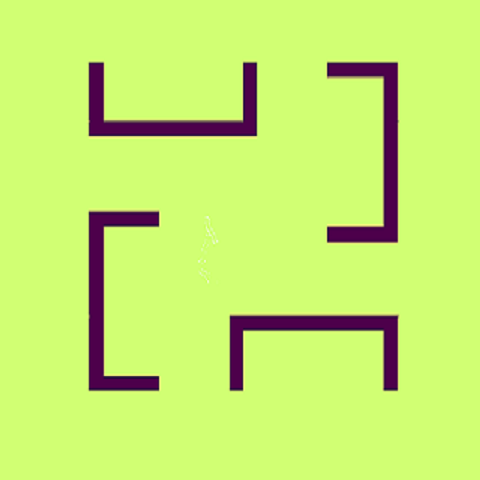}
    \caption{Firebreak Orientations.}
    \label{fig:orientations}
     \end{minipage}     

\end{figure}

To reduce the search space, the shape of firebreak cells proposed by \cite{martinez2022pattern} is used. Each figure corresponds to a rectangle-alike block consisting of 20 cells, each representing a firebreak. This firebreak pattern was selected because this study proves its effectiveness to contain the fire compared to other topologies of solutions, because its shape allows to enclose the fire inside it. An example of this configuration can be seen in Fig.\:\ref{fig:fconfig}. These sets of cells can be allocated in four different orientations, as shown Fig.\:\ref{fig:orientations}. For measuring its protective effect, random configurations of firebreaks with this shape are compared against randomly scattered cell configurations. 

\section{Wildfire simulation}  
To incorporate the intrinsic randomness of wildfires into the experiments, two sources of uncertainty are considered: ignition point and weather conditions.
As shown in Fig.\:\ref{fig:workScheme}, each fire starts with a random weather and a random ignition point. Three scenarios are created: \textit{M1}, a medium variability scenario in which the ignition points are located only in the central zone of the landscape, emulating a known distribution of ignition. The weather conditions consists of winds with random directions in any of the eight directions of the compass rose. \textit{M2} is a high variability scenario in which fires starts randomly in any point of the landscape, with the same weather conditions as in \textit{M1}. \textit{M3} is a medium variability scenario with random ignition points on any zone of the landscape but with real weather conditions of the zone, which have a more consistent distribution with less dispersion. Fig.\:\ref{fig:weather_dist} shows the distribution of these weather variables.

The duration of the fires is set in such a way that the entire forest is not burned but still burns a considerable portion of it. Thus, fires lasting 30 hours are simulated. After this, there is an assumption that rain or fire suppression will put out the fire. Weather conditions are assumed to remain constant throughout the 30 hours.
The wildfire simulation will use the software Cell2Fire, developed in \cite{pais2021cell2fire}, which simulates the fire propagation as a Cellular Automata.

\section{Results}\label{sec3}
The results section is divided into two subsections. First, both shapes of the solution proposed in Section \ref{secpattern} are evaluated by generating random solutions composed of firebreaks constructed from each type of solution. The results of this evaluation are used to construct metaheuristics, whose solutions are composed of the best type of topology resulting from the first subsection. The results of metaheuristics are then compared in different scenarios, according to the treatment percentage and the stochasticity level.

\subsection{Pattern Assessment}
\begin{figure}[ht]

        \centering
    \includegraphics[width=0.9\textwidth]{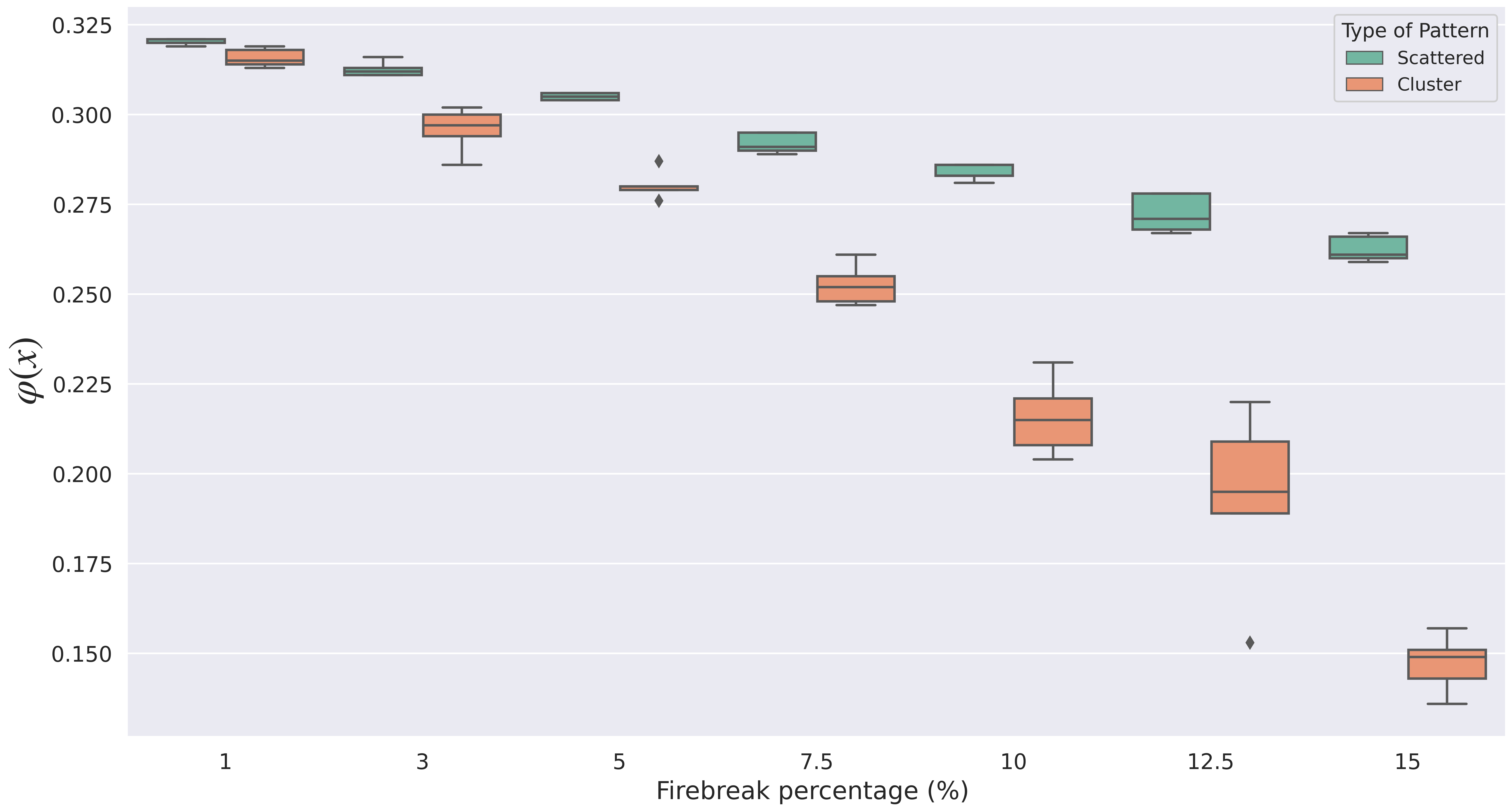}
    \caption{Comparison of pattern configuration.}
    \label{fig:pattern_assessment}
\end{figure}

\begin{figure}[!h]
    \begin{minipage}[t]{0.45\textwidth}
        \centering
        \includegraphics[width=0.8\textwidth]{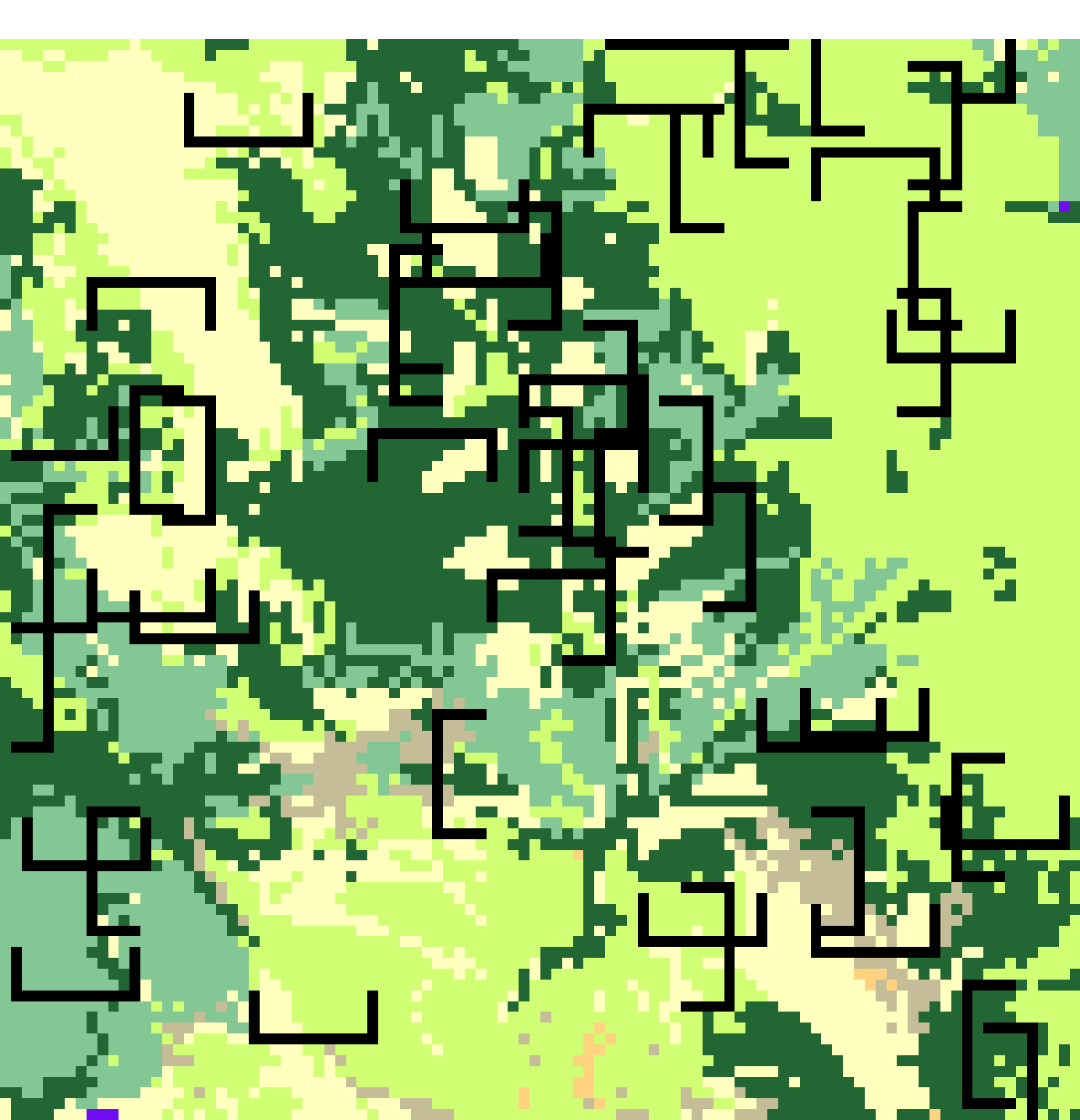}
        \caption{Cluster pattern for 10\% of firebreak.}
        \label{fig:cluster_example}
    \end{minipage}
    \hfill
    \begin{minipage}[t]{0.45\textwidth}
        \centering
    \includegraphics[width=0.8\textwidth]{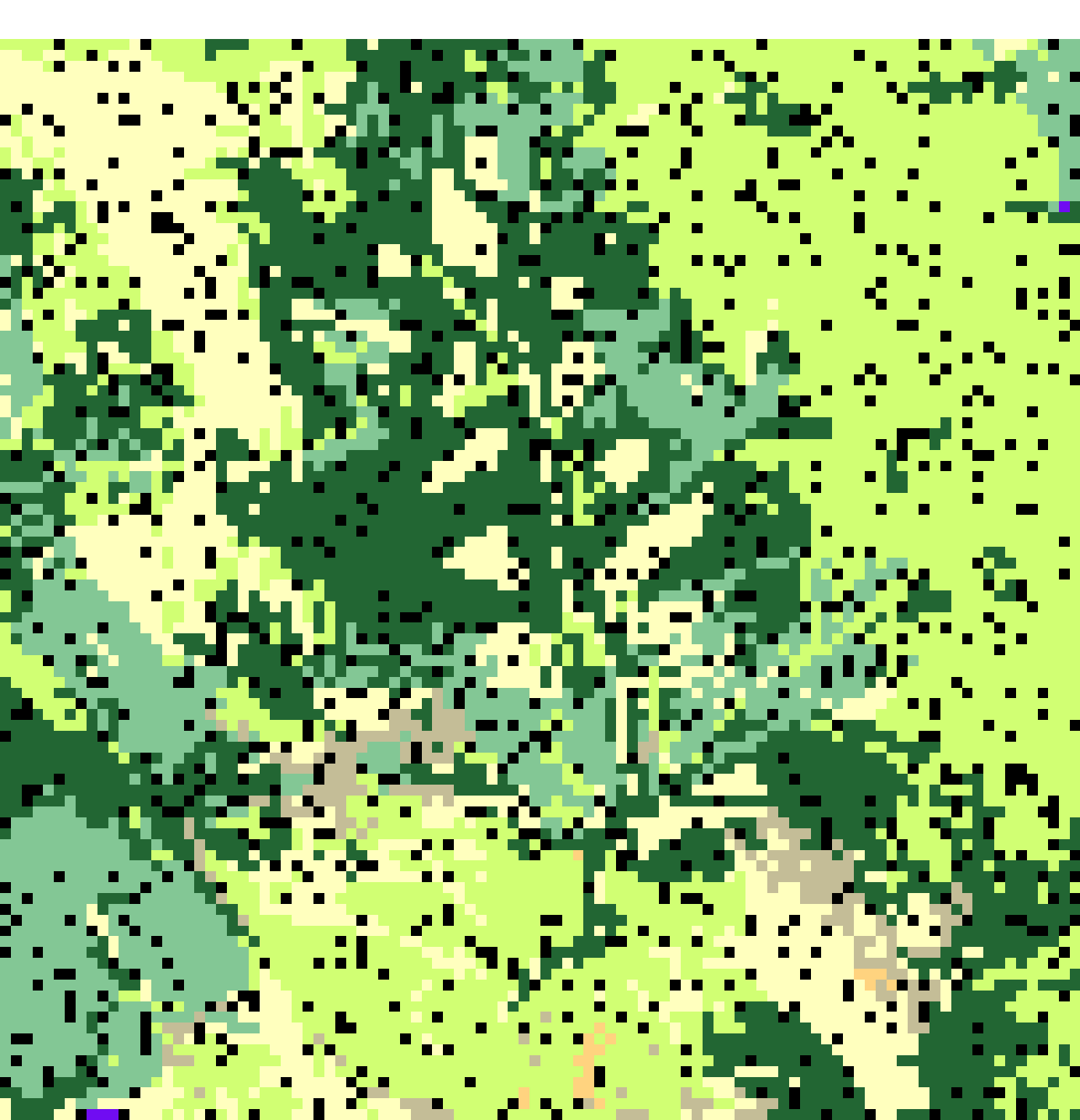}
    \caption{Scatter pattern for 10\% of firebreak.}
    \label{fig:scatter_example}
     \end{minipage}     

\end{figure}

The cluster pattern solutions have a better performance of the objective function $\varphi(x)$ than the scattered solutions. This can be seen in Fig.\:\ref{fig:pattern_assessment}, which shows that the cluster pattern solutions have an average of 24.3\% of burnt cells, while the scattered solutions have an average of 29.27\% of burnt cells.Detailed results can be found on Tables\:\ref{tab:clusterpattern}, \ref{tab:scatteredpattern}. 

Fig.\:\ref{fig:cluster_example} and Fig.\:\ref{fig:scatter_example} show an example of a solution with 10\% of firebreaks, the first constructed with clusters, while the latter is built through scattered firebreaks. The geometry of the cluster pattern allows the enclosure to fire more effectively due to its own topology. Also, each rectangle in the cluster pattern is more likely to overlap with other rectangles, unlike the scattered solutions. This overlap is shown to be effective in decreasing the forward fire spread in \cite{finney2001design}.

Therefore, all the solutions will be constructed using the cluster configuration from now on.

\subsection{Metaheuristics}

\subsubsection{\textit{M1} scenario}

\begin{figure}[ht]
    \centering
    \includegraphics[width=\textwidth]{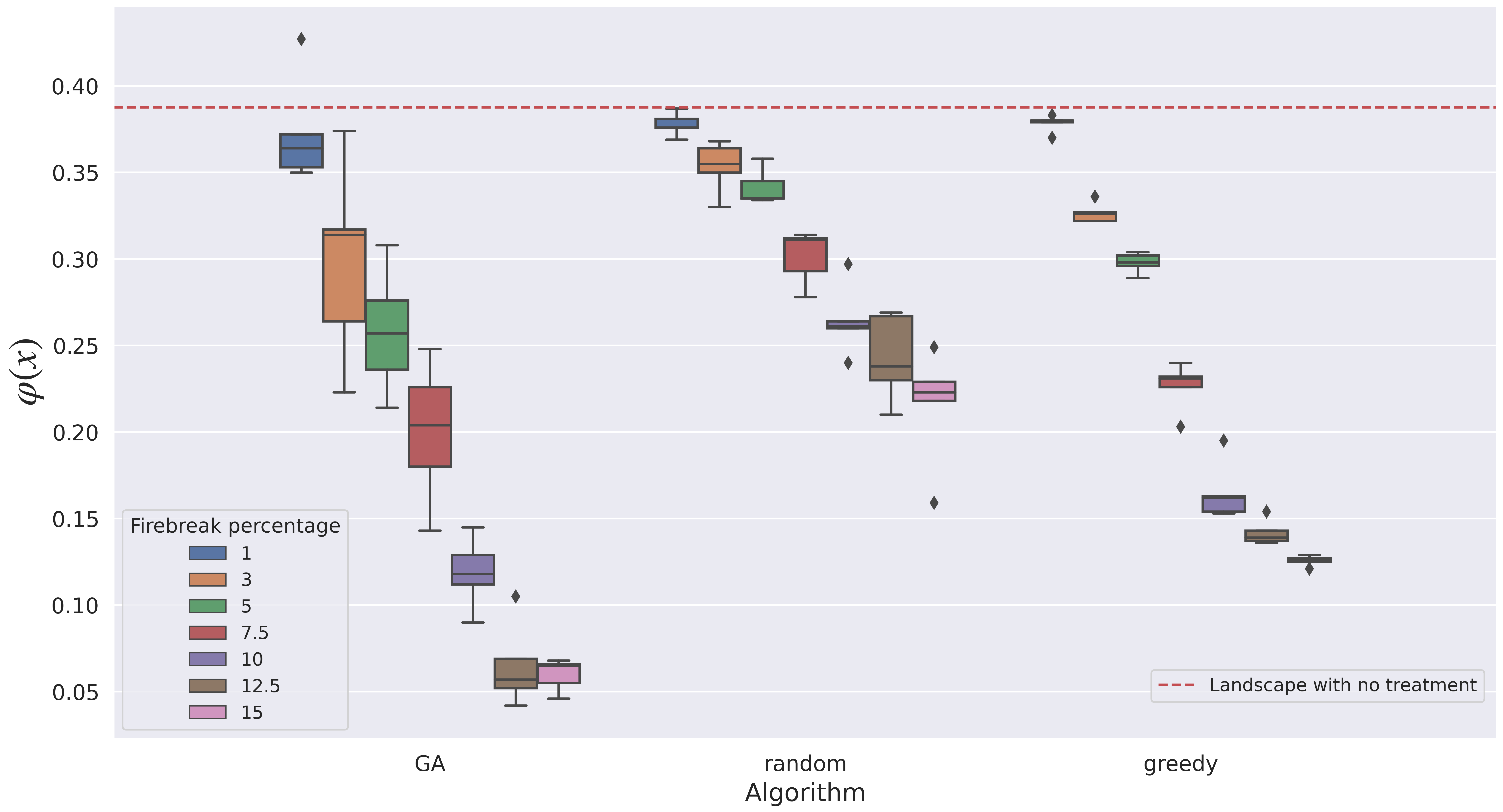}
    \caption{Burnt cells for \textit{M1} scenario.}
    \label{Fig:boxplotburntcellsquare}
\end{figure}

In \textit{M1}, GA shows the most robust performance at each percentage of firebreak. This can be seen in Fig.\:\ref{Fig:boxplotburntcellsquare}, which shows that GA has the lowest average burnt in each case. The biggest difference in the performance of GA is in the case with 12.5\% of the firebreak, in which it reaches an average of 6.5\% of burnt area, versus an average of 24.28\% and 14.18\% in random and greedy, respectively. The detailed results can be found in Tables\:\ref{tab:burntgam1}, \ref{tab:burntgraspm1}, \ref{tab:burntrandomm1}, \ref{tab:burntgreedym1}. 

\begin{figure}[ht]
        \centering
    \includegraphics[width=1.0\textwidth]{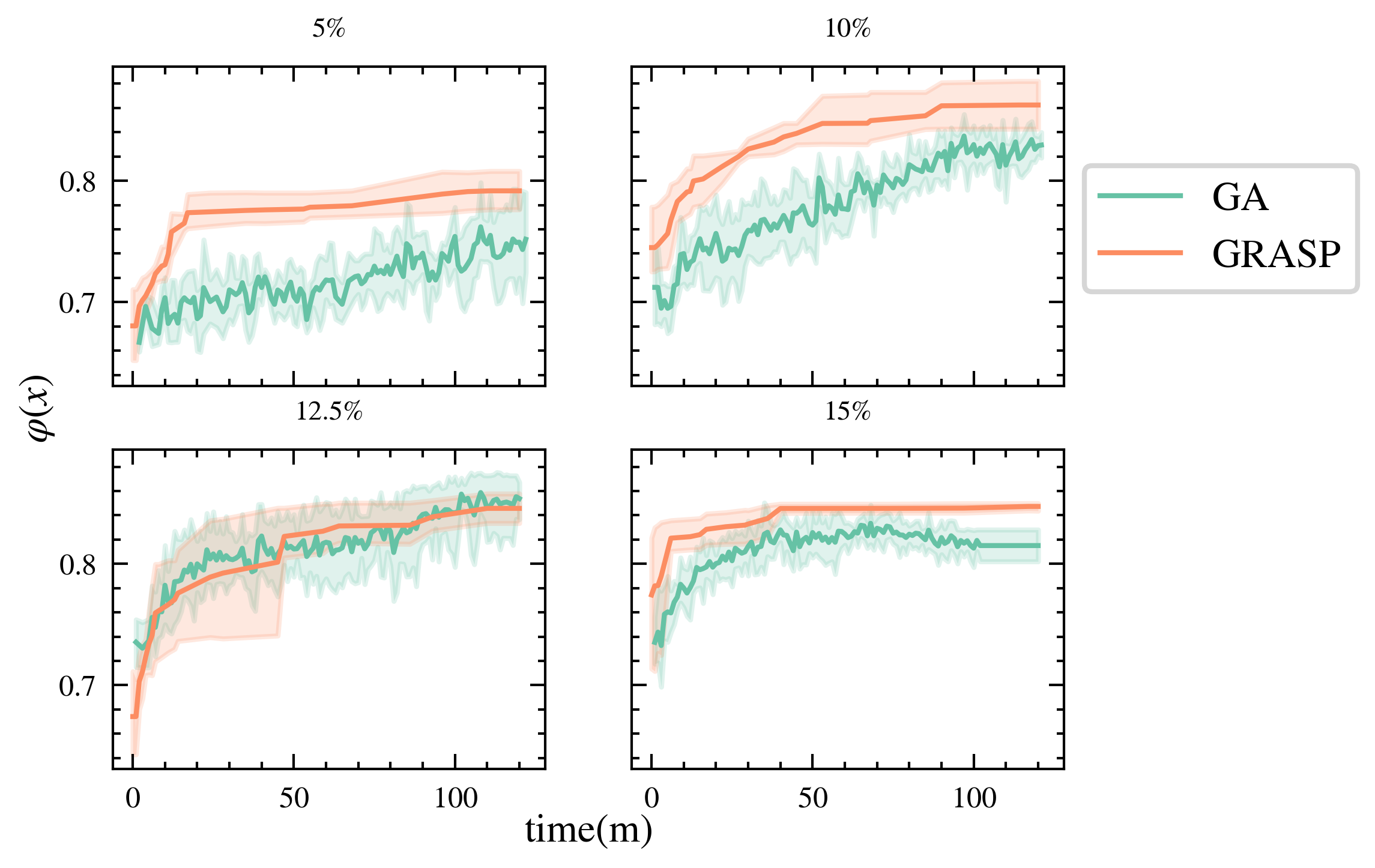}
    \caption{Evolution on time in \textit{M1} scenario for different firebreaks capability.}
    \label{fig:evolknown}
\end{figure}

In Fig.\:\ref{fig:evolknown}, the evolution of the performance along the execution can be seen. Each line represents the evolution of the global solution of the algorithm (in the case of GA, it shows the performance of the best individual of each generation). As it can be seen, GA improves its initial performance in the four cases shown. The improvement is small in the 5\% case, but it is large in the 10\% and 12.5\% cases. In the 15\% case, GA improves from the initial result, but afterward, its solution tends to decrease its performance. GRASP has a quick improvement in early iterations, but its rate of improvement decreases after that.

\subsubsection{\textit{M2} scenario}
\begin{figure}[ht]
    \centering
    \includegraphics[width=\textwidth]{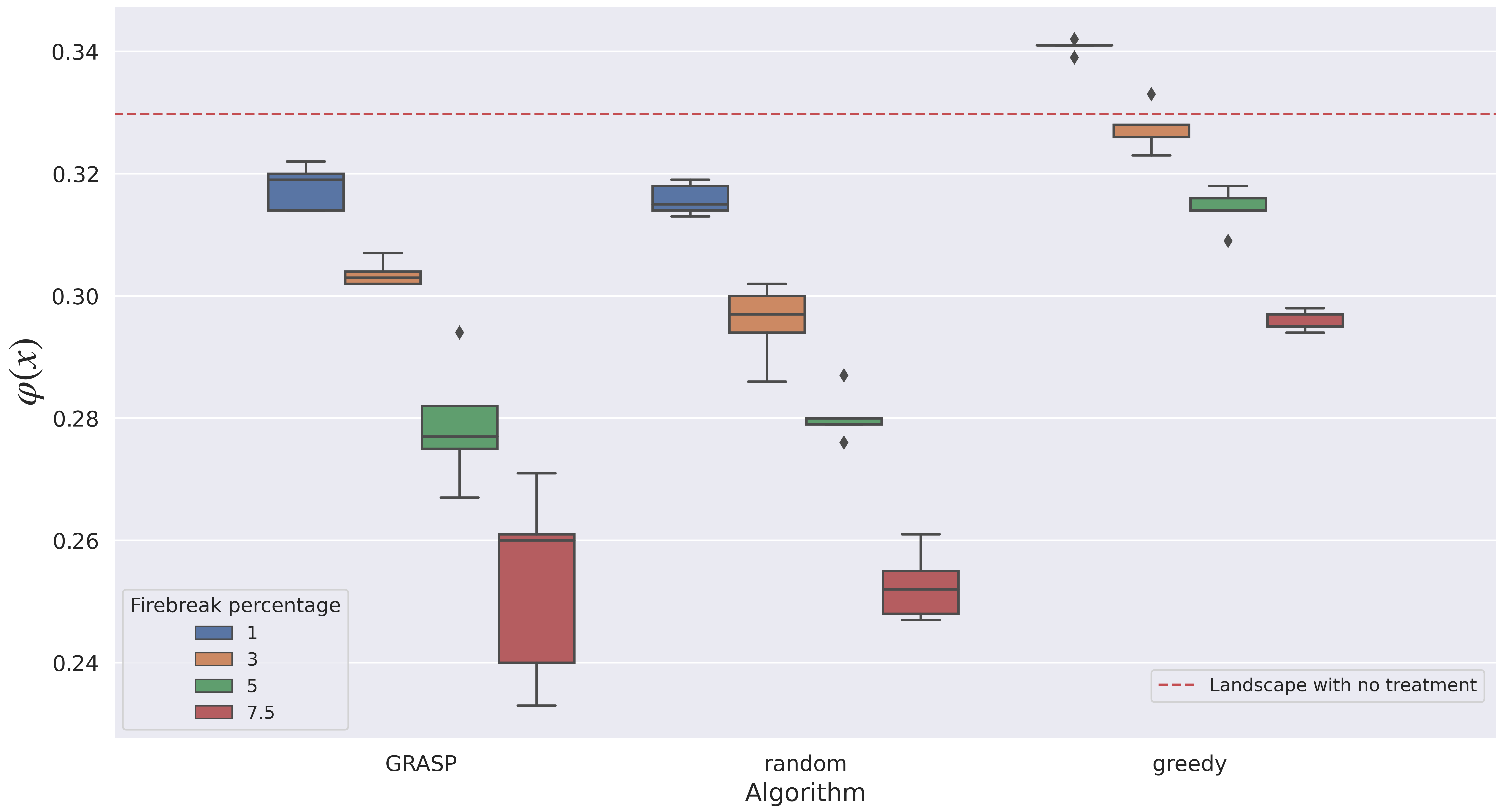}
    \caption{Burnt cells for low firebreak capacity in  \textit{M2} scenario.}
    \label{Fig:boxplotlowrandomburnt}
\end{figure}

In \textit{M2} scenario, the results are split into two sets. In low percentages of firebreak allocation, only GRASP has acceptable results, but its performance is similar or even worse than random solutions. Fig.\:\ref{Fig:boxplotlowrandomburnt} shows this.

\begin{figure}[ht]
    \centering
    \includegraphics[width=\textwidth]{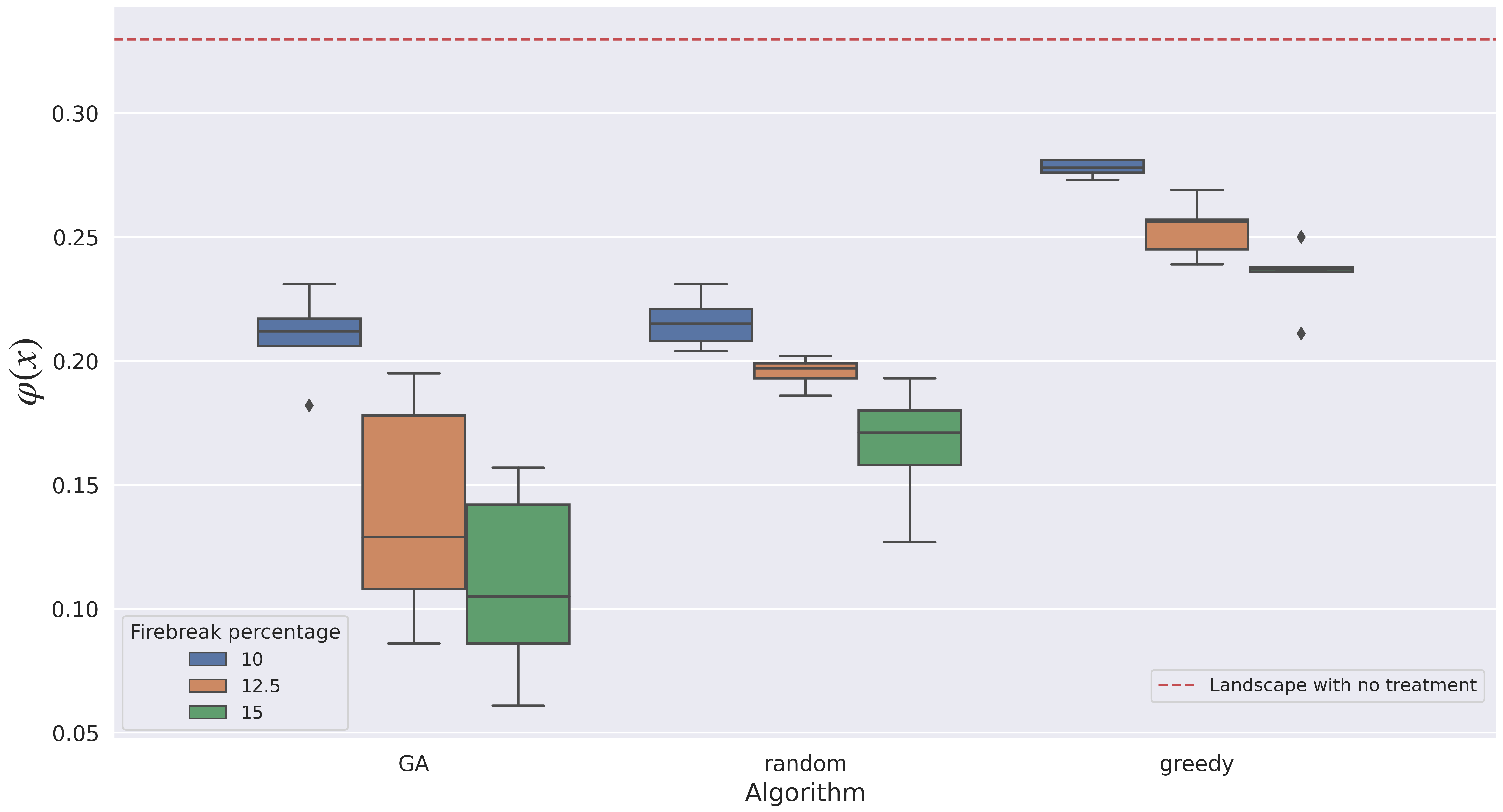}
    \caption{Burnt cells for high firebreak capacity in \textit{M2} scenario.}
    \label{Fig:boxplothighrandomburnt}
\end{figure}

Figure\:\ref{Fig:boxplothighrandomburnt} shows that at higher operational capability (more than 10\%), GA burns less area than any of the other methods (except on 10\% in which it is similar to random). For instance, at 15\% of firebreaks, GA reaches an average of 11.02\% of the area burnt. Meanwhile, random gets an average of 16.58\% and the greedy solution an average of 23.44\%.

Detailed results can be found on Tables\:\ref{tab:burntgam2}, \ref{tab:burntgraspm2}, \ref{tab:burntrandomm2}, \ref{tab:burntgreedym2}.

\begin{figure}[ht]
        \centering
    \includegraphics[width=1.0\textwidth]{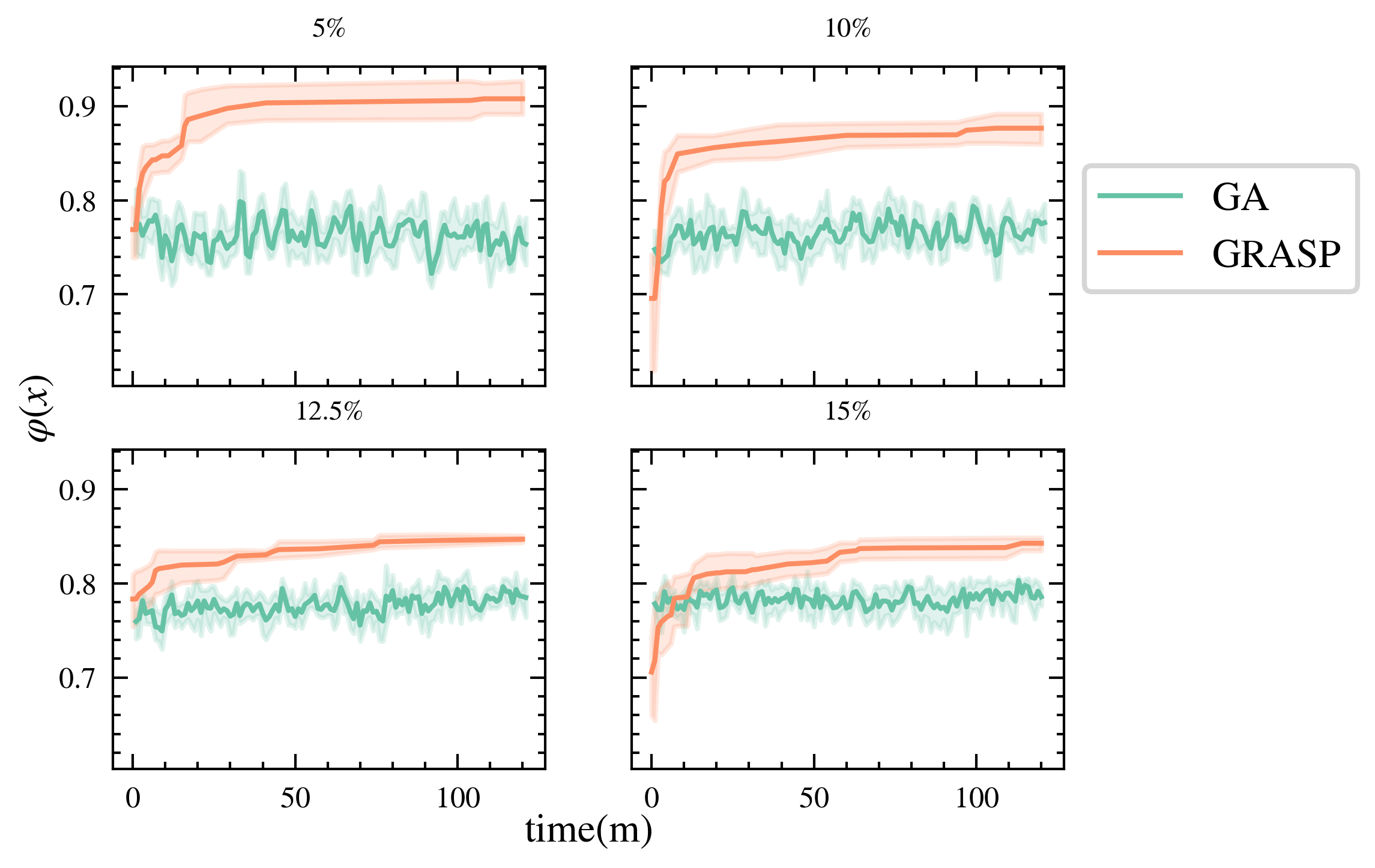}
    \caption{Evolution on time in \textit{M2} scenario for different firebreaks capability.}
    \label{fig:evolrandom}
\end{figure}

The results of the evolution of the performance in time (Fig.\:\ref{fig:evolknown}) show that, in general, GA is not able to improve from its initial results This means that the solutions generated by GA are not stable and can vary significantly over time. Meanwhile, GRASP is able to reach a high-performance solution quickly, and its performance only improves at a low rate in time. As mentioned in Section \ref{seccomp}, these results are highly noisy and do not represent the actual value of the objective function. Therefore, they should only be considered as a way to understand the inner workings of the algorithms. 

\subsubsection{\textit{M3} scenario}

\textit{M3} scenario has a similar behavior to \textit{M1}, this, a robust performance of GA against the other techniques. As Fig.\:\ref{Fig:boxplotburntcellsquare} shows, at each level of percentage GA overcomes the other techniques. As in \textit{M2}, the biggest difference in the performance is reached in the case with 12.5\% of the firebreak, in which GA burns an average of 28.08\% of the area, versus an average of 42.24\% and 62.02\% in random and greedy, respectively. The detailed results can be found in Tables\:\ref{tab:burntgam3}, \ref{tab:burntgraspm3}, \ref{tab:burntrandomm3}, \ref{tab:burntgreedym3}. 

\begin{figure}[ht]
    \centering
    \includegraphics[width=\textwidth]{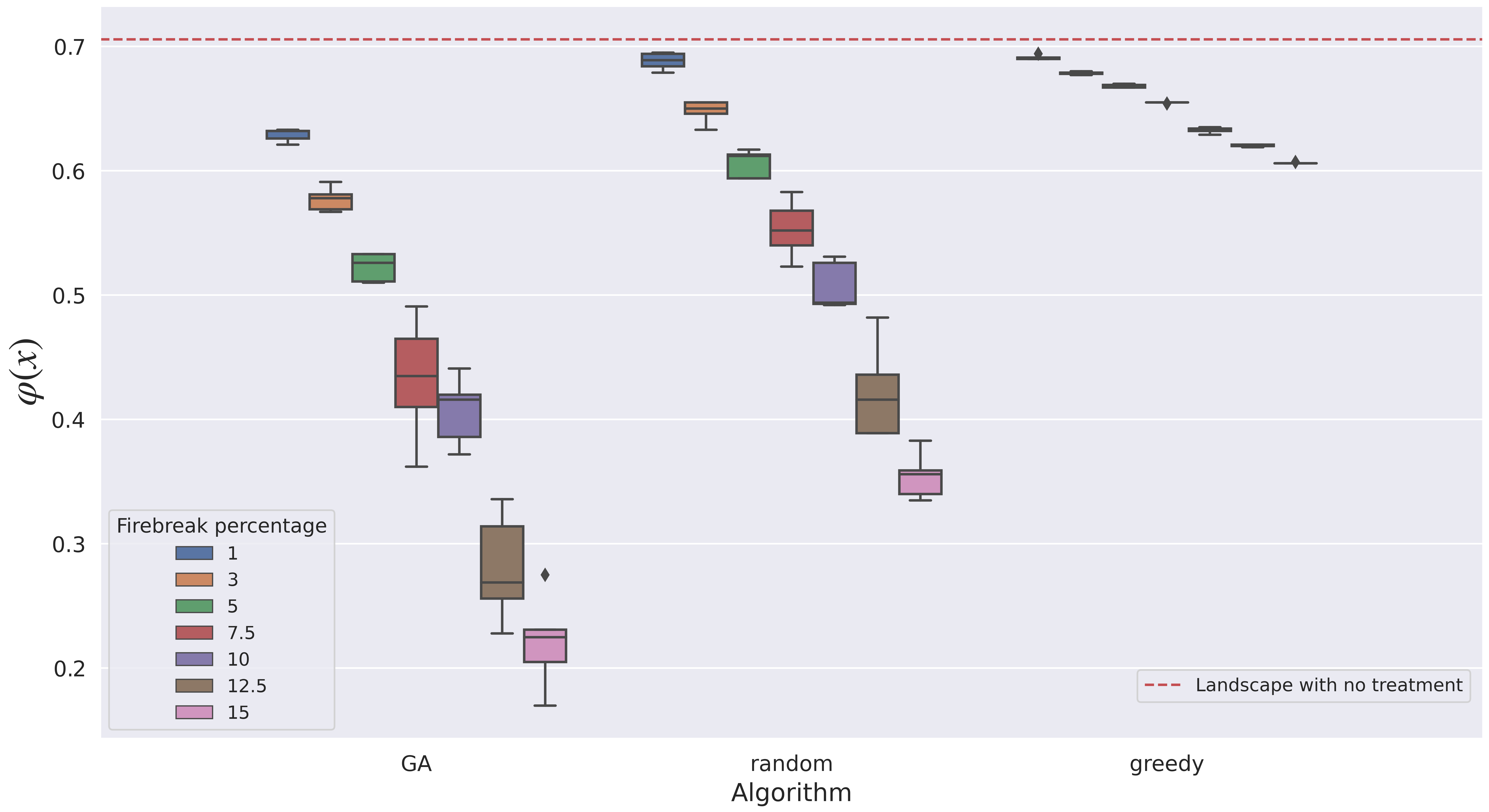}
    \caption{Burnt cells for \textit{M3} scenario.}
    \label{Fig:boxplotburntwind}
\end{figure}

\begin{figure}[ht]
        \centering
    \includegraphics[width=1.0\textwidth]{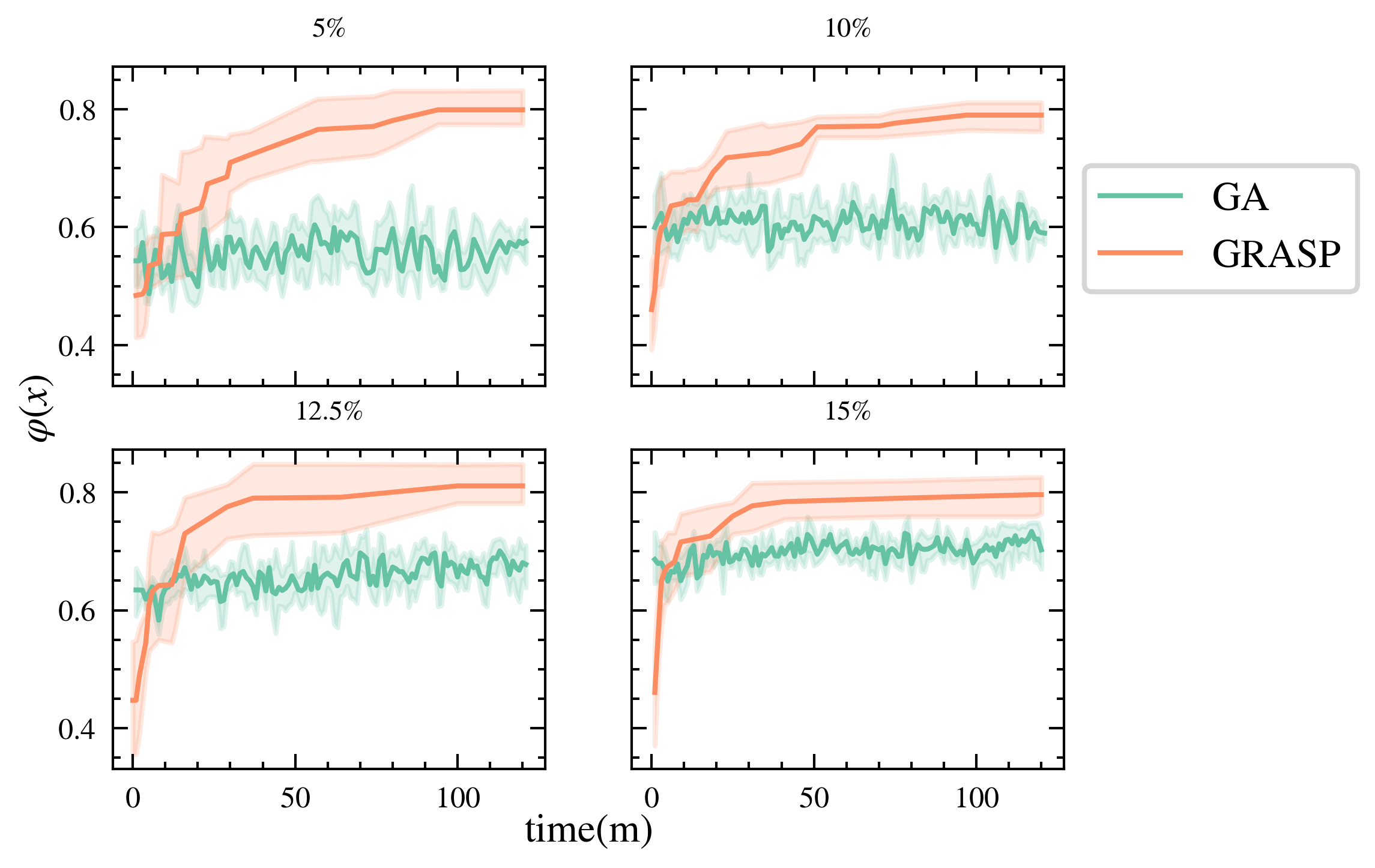}
    \caption{Evolution on time in \textit{M3} scenario for different firebreaks capability.}
    \label{fig:evolwind}
\end{figure}

The evolution of the performance in time for \textit{M3} (Fig.\:\ref{fig:evolwind}) shows that GA has an unstable behavior in the 5\% and 10\% cases, and a stable but with small improvement in time for the 12.5\% and 15\%. GRASP has a slow evolution in 5\%, 10\% and 12.5\% cases, compared to \textit{M1} and \textit{M2} scenarios, and it finds a good solution quickly in the 15\% case, with a small improvement rate after that.

\section{Discussion}\label{sec12}

GA performs better than GRASP because it ensures that the set of initial solutions evolves and does not return to an initial state or to a state similar to the one it had in the past, especially in cases with a high number of firebreaks. This is because the crossover between solutions ensures sufficient variability between solutions (and therefore less repetition of solutions), which prevents the algorithm from getting stuck in local minima. In contrast, GRASP tends to find solutions that are similar to the most frequently burned areas. This is because the greedy construction heuristic tends to favor these areas, and the local search heuristic is not able to escape from these. As a result, GRASP does not have enough variability in its solutions, which leads to worse results. The advantage of GRASP is that it has a high confidence value due to the high bias in the movement.

The algorithms seem to consistently improve as the number of firebreaks available increases. GA clearly outperforms GRASP. However, performance at low firebreak percentages must be considered. In such cases, in \textit{M2}, only GRASP performs similarly to the random methodology and beats the greedy method. In contrast, in \textit{M1} and \textit{M3}, GA outperforms both at any firebreak percentage, except for the case with 1\% in \textit{M1}, where there seems to be no clear advantage between the algorithm and the random solution. Therefore, the decision to include these metaheuristics should be considered only for cases with a moderately high operational capacity, considering the typical percentages treated in practice (see, e.g., \cite{oliveira2016assessing, jingan2005land}). This is especially important for scenarios such as \textit{M2}, where the algorithms do not perform well at low firebreak percentages. 

The performance of GA in \textit{M2} is affected by the stochasticity of the ignition and spread of fires  (represented by $\xi$ on Eq.\:(\ref{Eq:FMP})). This can be seen in the high standard deviation of the results, which makes it difficult to conclude with confidence whether GA is better than a random solution. The stochasticity is less powerful in \textit{M1} because the number of possible fires is reduced. This results in more deterministic solutions, which can be seen in the stability of GA towards the end of the execution on \textit{M1} (see Fig.\:\ref{fig:evolknown}) at the 15\% case, in which it clearly reaches a local optimal where the algorithm can no longer find a better or a worse solution. This can also be seen if we compare both \textit{M1} (Fig.\:\ref{fig:evolrandom}) and 
 \textit{M2} (Fig.\:\ref{fig:evolknown}). In the second, all the algorithms present a more progressive evolution over time, especially GA, which even seems to be able to obtain better solutions in hypothetical longer execution times for cases 10\% and 12.5\%, unlike \textit{M1}, where the algorithm is not able to converge to a stable solution. This implies that having a reduced set of possible scenarios allows the algorithms to converge to more reliable solutions. Considering this effect, it is crucial to reduce the stochasticity of this problem, using a method that approximates the behavior of the fires. Some possible ways to achieve this is to include probability maps that give a certain frequency of ignition for certain areas of the landscape, either considering potential risk or from historical ignitions \citep{carrasco2021exploring, 
 pais2021deep, de2001spatial}.
 Also, another way of reducing the stochasticity of the problem is to incorporate characteristic climatic conditions of the geographical area in order to obtain a certain range of wind speeds, temperature, relative humidity, etc  \citep{mcwethy2021broad, carmo2022climatology, giannaros2022meteorological}. This is done by the \textit{M3} scenario. From the Fig.\:\ref{fig:evolwind}, it can be seen that GA has a similar performance of the evolution as in \textit{M2} scenario. This suggests that the solutions in M3 would be of low quality due to the high variability of the conditions. However, GA is still able to outperform the other techniques at any percentage of firebreak, which indicates that despite having both stochastic ignitions and climatic conditions that cause an unstable behavior of the algorithm, the fact of having a limited distribution of climatic scenarios allows this metaheuristic to achieve better results than the other management techniques.

A suggested way to reduce the number of possible scenarios is to consider solving the problem for a subset of possible fires. This corresponds to solving this problem for only the $n$ worst fires, creating $\xi^n \subset \xi$. This would allow us to construct a landscape that is resistant to catastrophic wildfires but would still allow for smaller-scale wildfires. For example, in Canada, in the period 1957-1999, only the $3.1\%$ of the wildfires burnt $\sim$97\% of the total area burnt in that period \citep{stocks2002large}. In this case, $n$ could group fires with conditions similar to that $3.1\%$ instead of considering the whole set of conditions. 

\begin{figure}[ht]
    \centering
    \includegraphics[width=\textwidth]{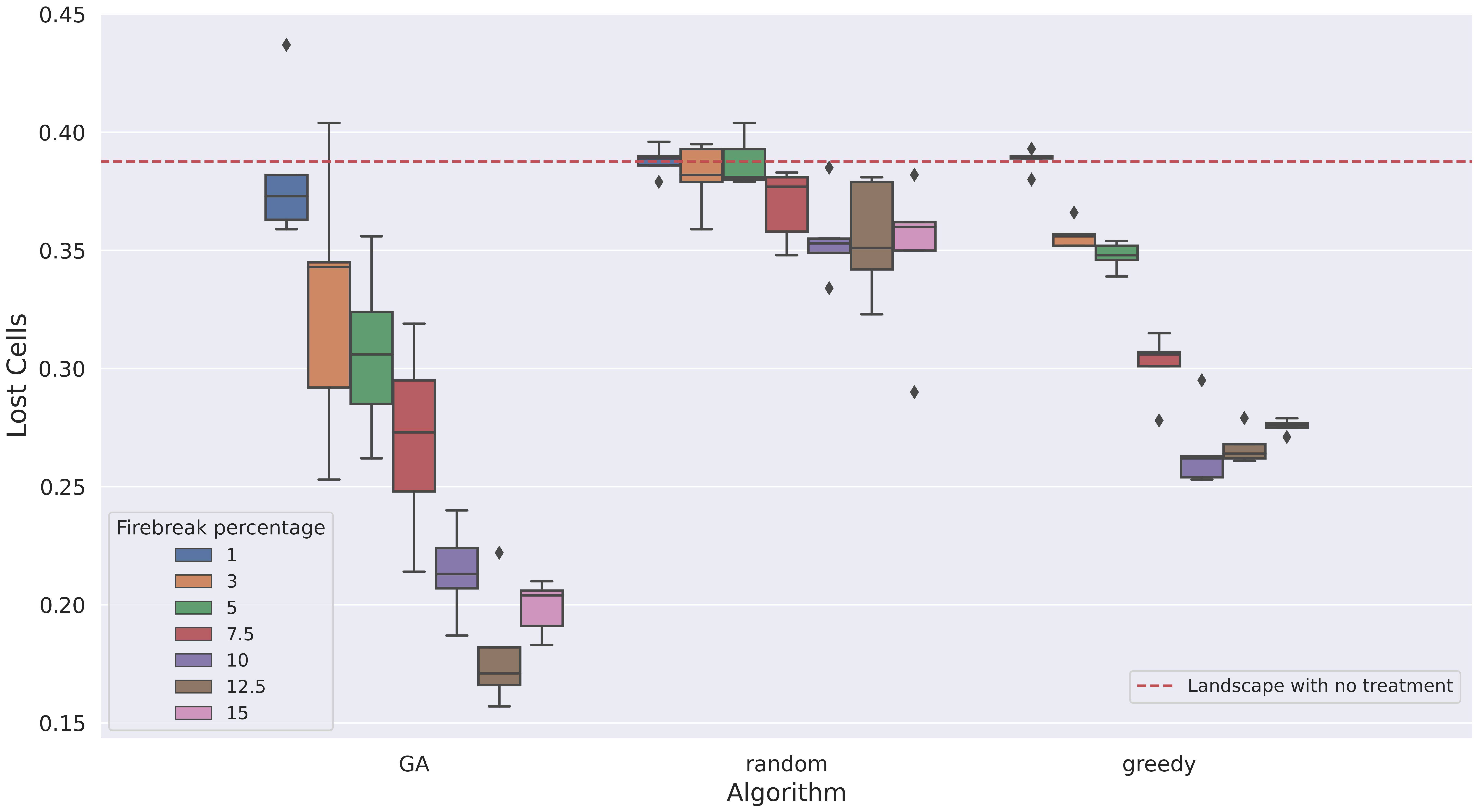}
    \caption{Lost cells according to Eq.\:(\ref{Eq:FMP2}) for \textit{M2} scenario.}
    \label{Fig:boxplotlostcellsquare}
\end{figure}

The placement of firebreaks can have negative implications \citep{stevens2016evaluating, scott2012short}. Therefore, it is important to consider the negative effects of firebreaks, as well as the negative effects of fires, when developing fire management strategies. One way to incorporate the negative effects of firebreaks is to modify the problem Eq.\:(\ref{Eq:FMP}) as follows:

\begin{equation} \label{Eq:FMP2}
    \min_{\vb{x} \in \Omega_\alpha} \varphi(\vb{x}):=\mathbb{E}_\mathcal{\xi} \left[ L \left(\vb{x},\xi \right) \right] + c({\vb{x}})
\end{equation}

where $c({\vb{x}})$ represents the losses associated with firebreak $\vb{x}$. If we assume that the entire value of the treated cell $\vb{x}$ is lost, then the solutions of the Eq.\:(\ref{Eq:FMP2}) for \textit{M1} would look as shown in Fig.\:\ref{Fig:boxplotlostcellsquare}. This figure shows that there is trade-off between placing more firebreaks to protect an area against fire and the loss associated with affecting the potential value at risk present in that area. For example, in the case of GA, if we compare the lost cells of 12.5\% and 15\% cases, we can see that a greater negative impact is being generated by allocating more treatments (if we only consider as $c({\vb{x}})$ the affectation to the potential value at risk, since the fire probably generates a greater impact if more aspects are considered) with the mere fact of including such treatments. Although the results of Eq.\:(\ref{Eq:FMP}) suggests that having more firebreaks means that the landscape is more protected against fires (in the case of GA, the average area burned is 6.5\% in the 12\% case versus 6\% in the 15\% case), this is not the case when we consider the loss of potential value at risk. In this new situation, the opposite occurs: the inclusion of more firebreaks only generates losses to the landscape by failing to protect more of the landscape than the one treated. 

Finally, it is important to consider that the shape of the objective function allows us to set an upper bound for this problem. This opens the door to the inclusion of other values in the objective function, for example, $c(\vb{x})$, mentioned above, associated with the loss due to the location of a firebreak. Other risk values related to wildfires, such as population density, the presence of critical infrastructure (e.g., power plants), and monetary loss from burned wood could also be added. The scalability of this problem also depends on the interaction with forestry experts and ecologists. This is necessary to correctly implement the metrics just mentioned and to generate model constraints that allow for more realistic solutions. These constraints could include connectivity constraints for firebreaks, feasibility constraints for locating them in certain places, the real effect of firebreaks on the spread of fire, and the eventual inclusion of other fuel treatment options such as pruning or thinning.

\section{Conclusions}\label{sec13}

This study proposed the application of simulation-based optimization (SbO) to solve the firebreak placement problem through the implementation and comparison of metaheuristics in order to reduce the area lost due to wildfires. The firebreak placement problem was formulated as a MIP that considers binary variables for the decision to place a firebreak in a given cell or not, in addition to an operational constraint that limits the number of cells to be treated.

For the approximation of the MIP objective value, we used a SbO approach based on the Cell2Fire fire simulator, together with the implementation of two metaheuristics: GA and GRASP.
We considered as a study area a heterogeneous portion of a Canadian forest in the Alberta region. Also, we generated three scenarios with medium and high stochasticity based on variables such as ignition point or weather conditions.

We compare the results for different amounts of treatment percentages. The first result showed that a clustered firebreaks provide significantly higher protection than scattered firebreaks. With this first result, we compare the metaheuristics considering clustered solutions and find that in a high stochasticity scenario, GRASP had the best performance but similar to a random solution at low treatment percentages, and GA performs well at high percentages. Meanwhile in two medium stochasticity scenarios, GA outperformed all other techniques. As a result, at $5\%$ treatment in medium stochasticity scenario according to the ignition zone, GA reduces the burned area in $\sim$13\% if it is compared with the untreated landscape and by $\sim$9\% if it is compared with a random solution. Therefore, metaheuristics, specifically the G.A. considering this study, are a valid method for solving the firebreak placement problem.

It is important to note that the results of SbO problems may contain considerable standard deviations. Therefore, there may be solutions that are superior on average but worse in certain ignition or weather scenarios. However, the scalability of the problem allows for solving this issue by increasing the execution time and thus providing more robust solutions.

The scope of this study allows for the inclusion of SbO through metaheuristics to solve the firebreak placement problem. While this study applied this solution to a given study area, it is possible to scale it to real landscapes with the inclusion of more realistic features (e.g., solution feasibility or treatment diversification). Since the problem is formulated as a MIP, the inclusion of such features can be done through the inclusion of constraints or modifications to the objective function. This requires the formulation of these characteristics with forestry experts in forest management, as well as government agencies, to consider budgetary, ecological, or legal constraints.

\backmatter

\section*{Declarations}

\bmhead{Conflict of interest}
The authors declare that they have no conflict of interest which is relevant to the content of this article.

\bmhead{Acknowledgments}
This project has received funding from the European Union’s Horizon 2020 research and innovation programme under grant agreement No 101037419 (\href{https://fire-res.eu}{FIRE-RES}) and the Complex Engineering Systems Institute (CONICYT-PIA-FB0816). The authors acknowledge the support of the Agencia Nacional de Investigaci\'on y Desarrollo (ANID), Chile, through project FONDEF  ID20I10137. \textbf{JC} acknowledges the support of the ANID, through funding Postdoctoral Fondecyt project No 3210311.

\clearpage

\begin{appendices}

\section{Metaheuristics description}\label{secA1}

    \begin{figure}[H]
    \centering
\usetikzlibrary{patterns}
\resizebox{0.8\textwidth}{!}{

\begin{tikzpicture}
\tikzstyle{startstop} = [rectangle, rounded corners, minimum width=3cm, minimum height=1cm,text centered, draw=black, fill=olive!5]
\tikzstyle{io} = [trapezium, trapezium left angle=70, trapezium right angle=110, minimum width=1cm, minimum height=1cm, text centered,text width=3cm,trapezium stretches=true, draw=black, fill=olive!5]
\tikzstyle{process} = [rectangle, minimum width=3cm, minimum height=1cm, text centered, draw=black, fill=olive!5]
\tikzstyle{decision} = [diamond, minimum width=3cm, minimum height=1cm, text centered, draw=black, fill=olive!5]
\tikzstyle{arrow} = [thick,->,>=Triangle]

\node (start) [startstop] {Start};
\node (pro1) [process, below of=start, yshift=-2cm] {Construction Phase};
\node (pro2) [process, right of=pro1, xshift=2.5cm] {Local Search};
\node (out1) [io, right of=pro2, xshift=2.5cm] {Local Optimum};
\node (pro3) [process, right of=out1, xshift=2.5cm] {Global Update};
\node (dec1) [decision, right of=pro3, xshift=2.5cm] {$runtime$\textgreater$T_{max}$?};
\node (out2) [io, above of=dec1,yshift=2cm] {Final Solution};
\node (stop) [startstop, left of=out2, xshift=-4cm] {End};

\draw [arrow] (start.south) -- node[anchor=west]  {$i=0, runtime=0$} (pro1);
\draw [arrow] (pro1) -- (pro2);
\draw [arrow] (pro2) -- (out1);
\draw [arrow] (out1) -- (pro3);
\draw [arrow] (pro3) -- (dec1);

\draw [arrow] (dec1) -- node[anchor=east] {yes} (out2);
\draw [arrow] (dec1.south) -- ++(0,-2)  |- ++(-7,0)  --node[pos=0,sloped,below] {no, $i=i+1$, $runtime=runtime+runtime_{i}$}  ++(-7,0)  --  (pro1) ;

\draw [arrow] (out2) -- (stop);

  \end{tikzpicture}}
   \caption[GRASP]{GRASP}
   \label{fig:graspall}
\end{figure}

\begin{figure}[h]
    \centering
    \includegraphics[width=0.8\textwidth]{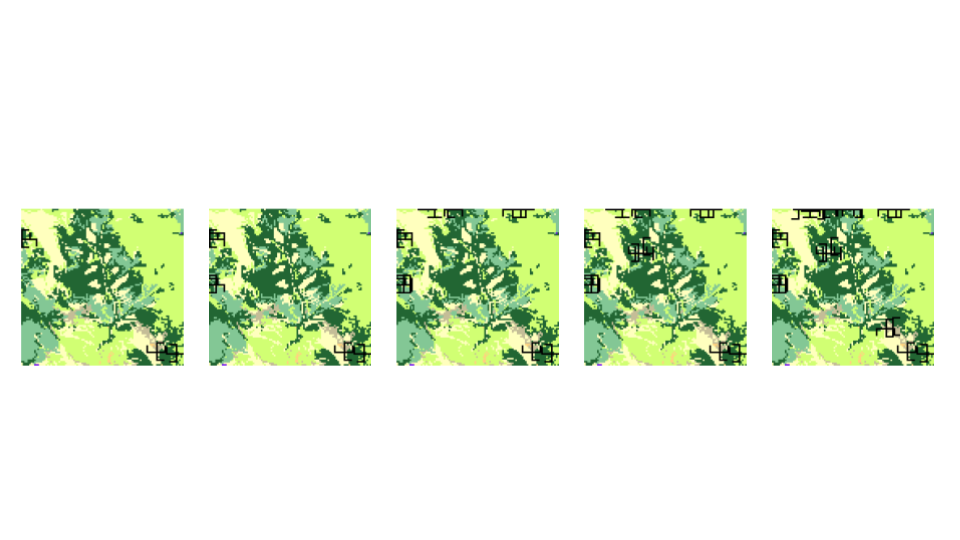}
    \caption{Solution construction for one iteration of GRASP. It can be seen (from left to right) the progressive allocation of firebreaks.}
    \label{fig:evolutiongrasp}
\end{figure}

    \begin{figure} [H]
    \centering
\usetikzlibrary{patterns}
\resizebox{\textwidth}{!}{

\begin{tikzpicture}
\tikzstyle{startstop} = [rectangle, rounded corners, minimum width=3cm, minimum height=1cm,text centered, draw=black, fill=olive!5]
\tikzstyle{io} = [trapezium, trapezium left angle=70, trapezium right angle=110, minimum width=1cm, minimum height=1cm, text centered,text width=3cm,trapezium stretches=true, draw=black, fill=olive!5]
\tikzstyle{process} = [rectangle, minimum width=3cm, minimum height=1cm, text centered, draw=black, fill=olive!5]
\tikzstyle{decision} = [diamond, minimum width=3cm, minimum height=1cm, text centered, draw=black, fill=olive!5]
\tikzstyle{arrow} = [thick,->,>=Triangle]

\node (start) [startstop] {Start};
\node (in1) [io, below of=start, yshift=-2cm] {Initial Generation};
\node (pro1) [process, right of=in1, xshift=2.5cm] {Simulation};
\node (pro2) [process, right of=pro1, xshift=2.5cm] {Selection};
\node (pro3) [process, right of=pro2, xshift=2.5cm] {Crossover};
\node (pro4) [process, right of=pro3, xshift=2.5cm] {Mutation};
\node (out1) [io, right of=pro4, xshift=2.5cm] {Generation $i$};
\node (dec1) [decision, right of=out1, xshift=2.5cm] {$runtime$\textgreater$T_{max}$?};
\node (out2) [io, below of=dec1,yshift=-2cm] {Final Solution};
\node (stop) [startstop, left of=out2, xshift=-4cm] {End};

\draw [arrow] (start.south) -- node[anchor=west]  {$i=0, runtime=0$} (in1);
\draw [arrow] (in1) -- (pro1);
\draw [arrow] (pro1) -- (pro2);
\draw [arrow] (pro2) -- (pro3);
\draw [arrow] (pro3) -- (pro4);
\draw [arrow] (pro4) -- (out1);
\draw [arrow] (out1) -- (dec1);
\draw [arrow] (dec1) -- node[anchor=east] {yes} (out2);
\draw [arrow] (dec1.north) -- ++(0,2)  |- ++(-7,0)  --node[pos=0,sloped,below] {no, $i=i+1$, $runtime=runtime+runtime_{i}$}  ++(-10.5,0)  --  (pro1) ;

\draw [arrow] (out2) -- (stop);

  \end{tikzpicture}}
   \caption[Genetic Algorithm]{Genetic Algorithm}
   \label{fig:allga}
\end{figure}

\begin{figure}[h]
    \centering
    \includegraphics[width=\textwidth]{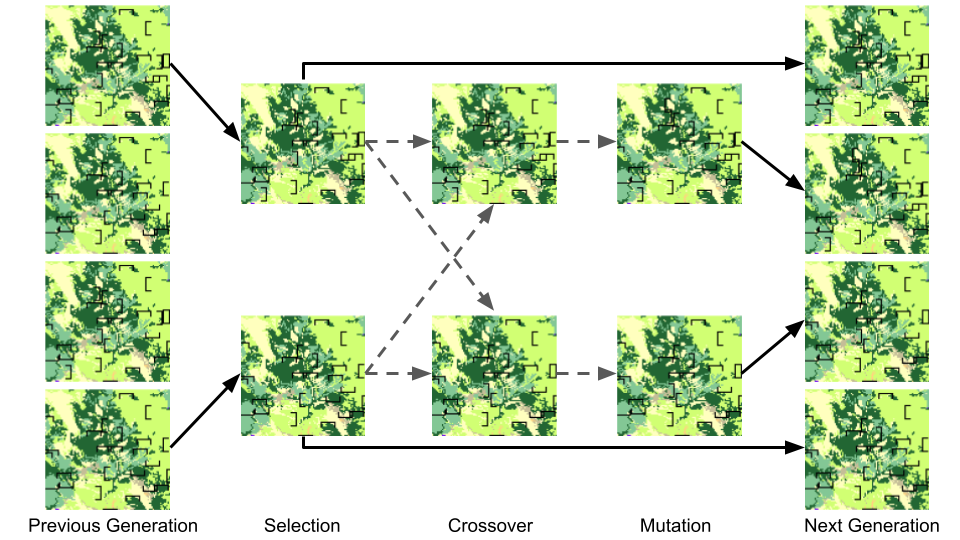}
    \caption{Construction of a generation of G.A. Here, some good candidates from the current generation are selected as members of future generation and also as parents, which will be combined between each other to generate offspring that will be mutated in order to get the second half of the future generation}
    \label{fig:evolutionga}
\end{figure}

\section{Instance Details}\label{secA2}

\begin{figure}[h]
    \centering
    \includegraphics[width=\textwidth]{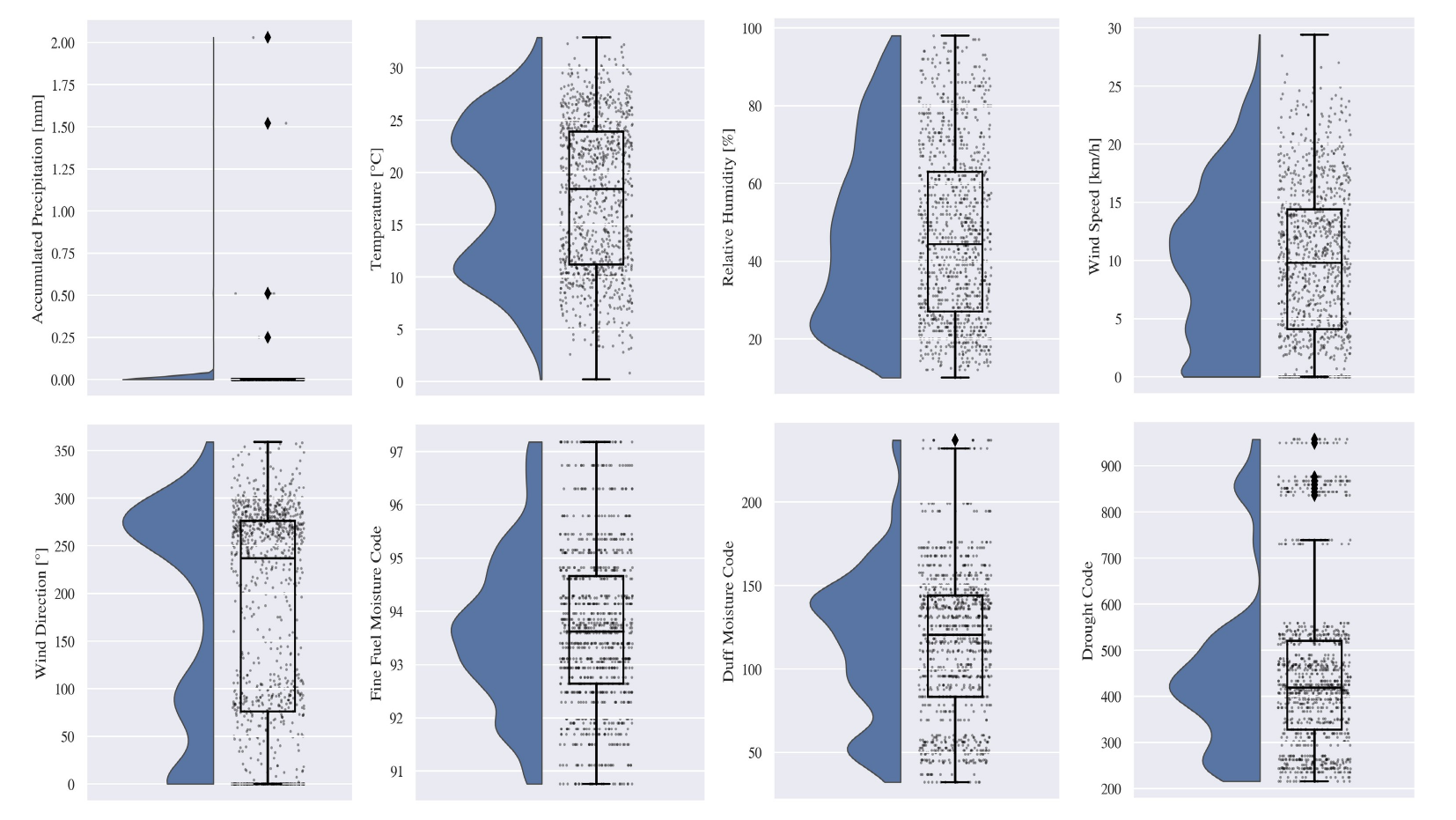}
    \caption{Weather distribution for \textit{M3}}
    \label{fig:weather_dist}
\end{figure}

\section{Results Details}\label{secA3}

\begin{table}[h]
    \centering
    \begin{tabular}{|c|c|c|c|c|c|c|c|}
    \hline
         \diagbox{N° Test}{Percentage}& 1\% & 3\% & 5\% & 7.5\% & 10\% & 12.5\% & 15\%   \\
         \hline
         1& 31.8 & 28.6 & 27.6 & 25.5 & 22.1 & 0.153 & 15.7 \\ 
          \hline
         2& 31.9 & 30 & 28 & 24.8 & 20.8 & 20.9 & 14.3 \\ 
            \hline
         3& 31.4 & 30.2 & 27.9 & 26.1 & 23.1 & 18.9 & 14.9 \\ 
            \hline
         4& 31.3 & 29.7 & 28 & 25.2 & 21.5 & 19.5 & 13.6 \\ 
            \hline
         5& 31.5 & 29.4 & 28.7 & 24.7 & 20.4 & 22 & 15.1 \\ 
         \hline
         \hline
         Mean& 31.58 & 29.58 & 28.04 & 25.26 & 21.58 & 19.32 & 14.72 \\ 
\hline
    \end{tabular}
    \caption{Burnt area of cluster pattern configurations}
    \label{tab:clusterpattern}
\end{table}

\begin{table}[h]
    \centering
    \begin{tabular}{|c|c|c|c|c|c|c|c|}
    \hline
         \diagbox{N° Test}{Percentage}& 1\% & 3\% & 5\% & 7.5\% & 10\% & 12.5\% & 15\%   \\
         \hline
         1& 32 & 31.1 & 30.4 & 28.9 & 28.1 & 27.8 & 26 \\ 
          \hline
         2& 32 & 31.6 & 30.6 & 29 & 28.3 & 26.8 & 26.1 \\ 
            \hline
         3& 32.1 & 31.3 & 30.5 & 29.5 & 28.6 & 27.8 & 26.7 \\ 
            \hline
         4& 32.1 & 31.1 & 30.6 & 29.5 & 28.3 & 26.7 & 25.9 \\ 
            \hline
         5& 31.9 & 31.2 & 30.4 & 29.1 & 28.6 & 27.1 & 26.6 \\ 
            \hline
            \hline
         Mean& 32.02 & 31.26 & 30.5 & 29.2 & 28.38 & 27.24 & 26.26 \\ 
\hline
    \end{tabular}
    \caption{Burnt area of scattered pattern configurations}
    \label{tab:scatteredpattern}
\end{table}

\begin{table}[]
    \centering
    \begin{tabular}{|c|c|c|c|c|c|c|c|c|}
    \hline
         \diagbox{N° Test}{Percentage}& 1\% & 3\% & 5\% & 7.5\% & 10\% & 12.5\% & 15\%   \\
         \hline
         1& 35.3 & 26.4 & 30.8 & 20.4 & 12.9 & 5.7 & 6.5 \\ 
          \hline
         2& 42.7 & 37.4 & 27.6 & 24.8 & 11.2 & 5.2 & 4.6 \\ 
            \hline
         3& 37.2 & 31.4 & 23.6 & 22.6 & 9   & 4.2 & 6.8 \\ 
            \hline
         4& 35   & 22.3 & 21.4 & 18   & 11.8 & 6.9 & 6.6 \\ 
            \hline
         5& 36.4 & 31.7 & 25.7 & 14.3 & 14.5 & 10.5 & 5.5 \\ 
            \hline
            \hline
         Mean& 37.32 & 29.84 & 25.82 & 20.02 & 11.88 & 6.5 & 6 \\ 
\hline
    \end{tabular}
    \caption{Burnt area of GA in \textit{M1} scenario}
    \label{tab:burntgam1}
\end{table}

\begin{table}[]
    \centering
    \begin{tabular}{|c|c|c|c|c|c|c|c|c|}
    \hline
         \diagbox{N° Test}{Percentage}& 1\% & 3\% & 5\% & 7.5\% & 10\% & 12.5\% & 15\%   \\
         \hline
         1& 37.6 & 28.3 & 32.9 & 23.7 & 22.1 & 15.9 & 15 \\ 
          \hline
         2& 38.5 & 35.3 & 30.4 & 26.7 & 18.5 & 17.5 & 10.2 \\ 
            \hline
         3& 37.7 & 32   & 29.1 & 20.4 & 21.8 & 19.2 & 11.5 \\ 
            \hline
         4& 37.1 & 31.5 & 28.7 & 26.9 & 18.8 & 24.2 & 16.7 \\ 
            \hline
         5& 37.4 & 34.9 & 32.4 & 20.1 & 16.3 & 15.1 & 15.9 \\ 
            \hline
            \hline
         Mean& 37.66 & 32.4 & 30.7 & 23.56 & 19.5 & 18.38 & 13.86 \\ 
\hline
    \end{tabular}
    \caption{Burnt area of GRASP in \textit{M1} scenario}
    \label{tab:burntgraspm1}
\end{table}

\begin{table}[]
    \centering
    \begin{tabular}{|c|c|c|c|c|c|c|c|c|}
    \hline
         \diagbox{N° Test}{Percentage}& 1\% & 3\% & 5\% & 7.5\% & 10\% & 12.5\% & 15\%   \\
         \hline
         1& 38.7 & 33   & 33.5 & 31.2 & 26   & 21 & 22.3 \\ 
          \hline
         2& 38.1 & 36.4 & 33.5 & 27.8 & 26.1 & 23.8 & 22.9 \\ 
            \hline
         3& 37.6 & 35   & 34.5 & 29.3 & 29.7 & 26.9 & 24.9 \\ 
            \hline
         4& 36.9 & 36.8 & 33.4 & 31.1 & 26.4 & 26.7 & 21.8 \\ 
            \hline
         5& 38.1 & 35.5 & 35.8 & 31.4 & 24   & 23   & 15.9 \\ 
            \hline
            \hline
         Mean& 37.88 & 35.34 & 34.14 & 30.16 & 26.44 & 24.28 & 21.56 \\ 
\hline
    \end{tabular}
    \caption{Burnt area of random solution in \textit{M1} scenario}
    \label{tab:burntrandomm1}
\end{table}

\begin{table}[]
    \centering
    \begin{tabular}{|c|c|c|c|c|c|c|c|c|}
    \hline
         \diagbox{N° Test}{Percentage}& 1\% & 3\% & 5\% & 7.5\% & 10\% & 12.5\% & 15\%   \\
         \hline
         1& 38   & 32.6 & 29.8 & 22.6 & 15.4 & 13.7 & 12.7 \\ 
          \hline
         2& 37   & 32.2 & 30.2 & 24   & 15.3 & 14.3 & 12.1 \\ 
            \hline
         3& 37.9 & 32.7 & 29.6 & 23.2 & 16.3 & 13.6 & 12.9 \\ 
            \hline
         4& 38.3 & 32.2 & 28.9 & 20.3 & 19.5 & 13.9 & 12.5 \\ 
            \hline
         5& 38   & 33.6 & 30.4 & 23.1 & 16.2 & 15.4 & 12.6 \\ 
            \hline
            \hline
         Mean& 37.84 & 32.66 & 29.78 & 22.64 & 16.54 & 14.18 & 12.56 \\ 
\hline
    \end{tabular}
    \caption{Burnt area of greedy solution in \textit{M1} scenario}
    \label{tab:burntgreedym1}
\end{table}

\begin{table}[]
    \centering
    \begin{tabular}{|c|c|c|c|c|c|c|c|c|}
    \hline
         \diagbox{N° Test}{Percentage}& 1\% & 3\% & 5\% & 7.5\% & 10\% & 12.5\% & 15\%   \\
         \hline
         1& 34.6 & 29.4 & 29.9 & 29.4 & 21.7 & 19.5 & 10.5 \\ 
          \hline
         2& 34.9 & 32.8 & 29 & 27 & 21.2 & 17.8 & 14.2 \\ 
            \hline
         3& 34.6 & 32.6 & 28.3 & 29.1 & 23.1 & 10.8 & 15.7 \\ 
            \hline
         4& 35 & 31.2 & 29.1 & 23.6 & 18.2 & 8.6 & 6.1 \\ 
            \hline
         5& 34.8 & 33.2 & 30.1 & 27.4 & 20.6 & 12.9 & 8.6 \\ 
            \hline
            \hline
         Mean& 34.78 & 31.84 & 29.28 & 27.3 & 20.96 & 13.92 & 11.02 \\ 
\hline
    \end{tabular}
    \caption{Burnt area of GA in \textit{M2} scenario}
    \label{tab:burntgam2}
\end{table}

\begin{table}[]
    \centering
    \begin{tabular}{|c|c|c|c|c|c|c|c|c|}
    \hline
         \diagbox{N° Test}{Percentage}& 1\% & 3\% & 5\% & 7.5\% & 10\% & 12.5\% & 15\%   \\
         \hline
         1& 32.2 & 30.2 & 26.7 & 26 & 24.3 & 23 & 19.8 \\ 
          \hline
         2& 32 & 30.2 & 29.4 & 26.1 & 24.5 & 22.3 & 21.7 \\ 
            \hline
         3& 31.4 & 30.7 & 27.7 & 24 & 24.8 & 21.3 & 16.2 \\ 
            \hline
         4& 31.9 & 30.4 & 28.2 & 27.1 & 24.3 & 20.5 & 21.3 \\ 
            \hline
         5& 31.4 & 30.3 & 27.5 & 23.3 & 25.2 & 17.8 & 21 \\ 
            \hline
            \hline
         Mean& 31.78 & 30.36 & 27.9 & 25.3 & 24.62 & 20.98 & 20 \\ 
\hline
    \end{tabular}
    \caption{Burnt area of GRASP in \textit{M2} scenario}
    \label{tab:burntgraspm2}
\end{table}

\begin{table}[]
    \centering
    \begin{tabular}{|c|c|c|c|c|c|c|c|c|}
    \hline
         \diagbox{N° Test}{Percentage}& 1\% & 3\% & 5\% & 7.5\% & 10\% & 12.5\% & 15\%   \\
         \hline
         1& 31.8 & 28.6 & 27.6 & 25.5 & 22.1 & 19.3 & 15.8 \\ 
          \hline
         2& 31.9 & 30 & 28 & 24.8 & 20.8 & 18.6 & 17.1 \\ 
            \hline
         3& 31.4 & 30.2 & 27.9 & 26.1 & 23.1 & 20.2 & 19.3 \\ 
            \hline
         4& 31.3 & 29.7 & 28 & 25.2 & 21.5 & 19.9 & 18 \\ 
            \hline
         5& 31.5 & 29.4 & 28.7 & 24.7 & 20.4 & 19.7 & 12.7 \\ 
            \hline
            \hline
         Mean& 31.58 & 29.58 & 28.04 & 25.26 & 21.58 & 19.54 & 16.58 \\ 
\hline
    \end{tabular}
    \caption{Burnt area of random solution in \textit{M2} scenario}
    \label{tab:burntrandomm2}
\end{table}

\begin{table}[]
    \centering
    \begin{tabular}{|c|c|c|c|c|c|c|c|c|}
    \hline
         \diagbox{N° Test}{Percentage}& 1\% & 3\% & 5\% & 7.5\% & 10\% & 12.5\% & 15\%   \\
         \hline
         1& 34.2 & 32.8 & 31.4 & 29.5 & 27.3 & 24.5 & 23.8 \\ 
          \hline
         2& 33.9 & 32.3 & 30.9 & 29.5 & 27.8 & 25.7 & 23.7 \\ 
            \hline
         3& 34.1 & 32.6 & 31.4 & 29.8 & 28.1 & 26.9 & 25 \\ 
            \hline
         4& 34.1 & 33.3 & 31.8 & 29.7 & 28.1 & 25.6 & 23.6 \\ 
            \hline
         5& 34.1 & 32.8 & 31.6 & 29.4 & 27.6 & 23.9 & 21.1 \\ 
            \hline
            \hline
         Mean& 34.08 & 32.76 & 31.42 & 29.58 & 27.78 & 25.32 & 23.44 \\ 
\hline
    \end{tabular}
    \caption{Burnt area of greedy solution in \textit{M2} scenario}
    \label{tab:burntgreedym2}
\end{table}

\begin{table}[]
    \centering
    \begin{tabular}{|c|c|c|c|c|c|c|c|c|}
    \hline
         \diagbox{N° Test}{Percentage}& 1\% & 3\% & 5\% & 7.5\% & 10\% & 12.5\% & 15\%   \\
         \hline
         1& 62.6 & 56.7 & 52.6 & 43.5 & 41.6 & 26.9 & 20.5 \\ 
          \hline
         2& 63.3 & 59.1 & 53.3 & 49.1 & 38.6 & 25.6 & 17 \\ 
            \hline
         3& 62.1 & 57.8 & 53.3 & 36.2 & 37.2 & 22.8 & 27.5 \\ 
            \hline
         4& 63.2 & 56.9 & 51.1 & 46.5 & 44.1 & 31.5 & 22.5 \\ 
            \hline
         5& 63.2 & 58.1 & 51 & 41 & 42   & 33.6 & 23.1 \\ 
            \hline
            \hline
         Mean& 62.88 & 57.72 & 52.26 & 43.26 & 40.7 & 28.08 & 22.12 \\ 
\hline
    \end{tabular}
    \caption{Burnt area of GA in \textit{M3} scenario}
    \label{tab:burntgam3}
\end{table}

\begin{table}[]
    \centering
    \begin{tabular}{|c|c|c|c|c|c|c|c|c|}
    \hline
         \diagbox{N° Test}{Percentage}& 1\% & 3\% & 5\% & 7.5\% & 10\% & 12.5\% & 15\%   \\
         \hline
         1& 69.2 & 63.4 & 62    & 60.2 & 53.1 & 55.4 & 46.1 \\ 
          \hline
         2& 69.4 & 65.6 & 64.8  & 58.4 & 53.9 & 53.4 & 47.5 \\ 
            \hline
         3& 69.1 & 66.2 & 61.3 & 60.5 & 53.8 & 45 & 47.4 \\ 
            \hline
         4& 69.1 & 64.8 & 62.7  & 59.3 & 58.4 & 51.8 & 46.4 \\ 
            \hline
         5& 69.1 & 65.7 & 63.1  & 58.7 & 57.4 & 54.4 & 42.1 \\ 
            \hline
            \hline
         Mean& 69.18 & 65.14 & 62.78 & 59.42 & 55.32 & 52 & 45.9 \\ 
\hline
    \end{tabular}
    \caption{Burnt area of GRASP in \textit{M3} scenario}
    \label{tab:burntgraspm3}
\end{table}

\begin{table}[]
    \centering
    \begin{tabular}{|c|c|c|c|c|c|c|c|c|}
    \hline
         \diagbox{N° Test}{Percentage}& 1\% & 3\% & 5\% & 7.5\% & 10\% & 12.5\% & 15\%   \\
         \hline
         1& 68.4 & 65 & 59.4 & 54 & 49.2 & 41.6 & 35.6 \\ 
          \hline
         2& 67.9 & 65.5 & 61.7 & 55.2 & 49.3 & 38.9 & 38.3 \\ 
            \hline
         3& 69.5 & 64.6 & 59.4 & 58.3 & 49.4 & 38.9 & 33.5 \\ 
            \hline
         4& 69.4 & 63.3 & 61.2 & 52.3 & 52.6 & 43.6 & 34 \\ 
            \hline
         5& 68.9 & 65.5 & 61.3 & 56.8 & 53.1 & 48.2 & 35.9 \\ 
            \hline
            \hline
         Mean& 68.82 & 64.78 & 60.6 & 55.32 & 50.72 & 42.24 & 35.46 \\ 
\hline
    \end{tabular}
    \caption{Burnt area of random solution in \textit{M3} scenario}
    \label{tab:burntrandomm3}
\end{table}

\begin{table}[]
    \centering
    \begin{tabular}{|c|c|c|c|c|c|c|c|c|}
    \hline
         \diagbox{N° Test}{Percentage}& 1\% & 3\% & 5\% & 7.5\% & 10\% & 12.5\% & 15\%   \\
         \hline
         1& 69.1 & 67.9 & 66.9 & 65.4 & 63.2 & 62.1 & 60.6 \\ 
          \hline
         2& 69.1 & 67.8 & 67   & 65.5 & 63.4 & 62 & 60.6 \\ 
            \hline
         3& 69   & 67.7 & 66.7 & 65.5 & 63.5 & 62 & 60.6 \\ 
            \hline
         4& 69.4 & 67.9 & 66.7 & 65.5 & 63.4 & 61.9 & 60.7 \\ 
            \hline
         5& 69   & 68   & 66.9 & 65.5 & 62.9 & 62.1 & 60.6 \\ 
            \hline
            \hline
         Mean& 69.12 & 67.86 & 66.84 & 65.48 & 63.28 & 62.02 & 60.62 \\ 
\hline
    \end{tabular}
    \caption{Burnt area of greedy solution in \textit{M3} scenario}
    \label{tab:burntgreedym3}
\end{table}

\begin{table}[]
    \centering
    \begin{tabular}{|c|c|c|c|c|c|c|c|c|}
    \hline
         \diagbox{N° Test}{Percentage}& 1\% & 3\% & 5\% & 7.5\% & 10\% & 12.5\% & 15\%   \\
         \hline
         1& 36.3 & 29.2 & 35.6 & 27.3 & 22.4 & 17.1 & 20.4 \\ 
          \hline
         2& 43.7 & 40.4 & 32.4 & 31.9 & 20.7 & 16.6 & 18.3 \\ 
            \hline
         3& 38.2 & 34.3 & 28.5 & 29.5 & 18.7 & 15.7 & 21 \\ 
            \hline
         4& 35.9 & 25.3 & 26.2 & 24.8 & 21.3 & 18.2 & 20.6 \\ 
            \hline
         5& 37.3 & 34.5 & 30.6 & 21.4 & 24   & 22.2 & 19.1 \\ 
            \hline
            \hline
         Mean& 38.28 & 32.74 & 30.66 & 26.98 & 21.42 & 17.96 & 19.88 \\ 
\hline
    \end{tabular}
    \caption{Lost area of GA in \textit{M1} scenario}
    \label{tab:lostgam1}
\end{table}

\begin{table}[]
    \centering
    \begin{tabular}{|c|c|c|c|c|c|c|c|c|}
    \hline
         \diagbox{N° Test}{Percentage}& 1\% & 3\% & 5\% & 7.5\% & 10\% & 12.5\% & 15\%   \\
         \hline
         1& 38.5 & 31.2 & 37.9 & 31   & 32   & 28.3 & 29.9 \\ 
          \hline
         2& 39.4 & 38.2 & 35.2 & 34   & 28.5 & 29.9 & 25.1 \\ 
            \hline
         3& 38.5 & 34.9 & 34   & 27.8 & 31.7 & 31.6 & 26.5 \\ 
            \hline
         4& 38   & 34.5 & 33.7 & 34.4 & 28.7 & 36.6 & 31.6 \\ 
            \hline
         5& 38.3 & 37.8 & 37.4 & 27.6 & 26.3 & 27.5 & 30.8 \\ 
            \hline
            \hline
         Mean& 38.54 & 35.32 & 35.64 & 30.96 & 29.44 & 30.78 & 28.78 \\ 
\hline
    \end{tabular}
    \caption{Lost area of GRASP in \textit{M1} scenario}
    \label{tab:lostgraspm1}
\end{table}

\begin{table}[]
    \centering
    \begin{tabular}{|c|c|c|c|c|c|c|c|c|}
    \hline
         \diagbox{N° Test}{Percentage}& 1\% & 3\% & 5\% & 7.5\% & 10\% & 12.5\% & 15\%   \\
         \hline
         1& 39.6 & 35.9 & 38.1 & 37.7 & 34.9 & 32.3 & 36 \\ 
          \hline
         2& 39   & 39.3 & 38   & 34.8 & 35.5 & 35.1 & 36.2 \\ 
            \hline
         3& 38.6 & 37.9 & 39.3 & 35.8 & 38.5 & 37.9 & 38.2 \\ 
            \hline
         4& 37.9 & 39.5 & 37.9 & 38.1 & 35.3 & 38.1 & 35 \\ 
            \hline
         5& 38.9 & 38.2 & 40.4 & 38.3 & 33.4 & 34.2 & 29 \\ 
            \hline
            \hline
         Mean& 38.8 & 38.18 & 38.74 & 36.94 & 35.52 & 35.52 & 34.88 \\ 
\hline
    \end{tabular}
    \caption{Lost area of random solution in \textit{M1} scenario}
    \label{tab:lostrandomm1}
\end{table}

\begin{table}[]
    \centering
    \begin{tabular}{|c|c|c|c|c|c|c|c|c|}
    \hline
         \diagbox{N° Test}{Percentage}& 1\% & 3\% & 5\% & 7.5\% & 10\% & 12.5\% & 15\%   \\
         \hline
         1& 39 & 35.6 & 34.8 & 30.1 & 25.4 & 26.2 & 27.7 \\ 
          \hline
         2& 38   & 35.2 & 35.2   & 31.5 & 25.3 & 26.8 & 27.1 \\ 
            \hline
         3& 38.9 & 35.7 & 34.6 & 30.7 & 26.3 & 26.1 & 27.9 \\ 
            \hline
         4& 39.3 & 35.2 & 33.9 & 27.8 & 29.5 & 26.4 & 27.5 \\ 
            \hline
         5& 39 & 36.6 & 35.4 & 30.6 & 26.2 & 27.9 & 27.6 \\ 
            \hline
            \hline
         Mean& 38.84 & 35.66 & 34.78 & 30.14 & 26.54 & 26.68 & 27.56 \\ 
\hline
    \end{tabular}
    \caption{Lost area of greedy solution in \textit{M1} scenario}
    \label{tab:lostgreedym1}
\end{table}




\end{appendices}

\clearpage
\bibliography{sn-bibliography}


\end{document}